\pdfoutput=1

\documentclass[11pt]{article}

\usepackage{acl}

\usepackage{times}
\usepackage{latexsym}

\usepackage[T1]{fontenc}

\usepackage[utf8]{inputenc}
\usepackage{hyperref}       
\usepackage{url}            
\usepackage{booktabs}       
\usepackage{amsfonts}       
\usepackage{nicefrac}       
\usepackage{microtype}      
\usepackage{xcolor}         
\usepackage{times}
\usepackage{latexsym}
\usepackage{graphicx}
\usepackage{subcaption}
\usepackage[ruled,vlined]{algorithm2e}
\usepackage{multirow}
\usepackage{tabularx}
\usepackage{amsmath}
\usepackage{bbold}
\usepackage{mathtools}
\usepackage{xcolor}
\usepackage{wrapfig}
\usepackage{tcolorbox}

\usepackage{hyperref}
\usepackage{url}
\usepackage{inconsolata}
\usepackage{amsmath}
\usepackage{graphicx}
\usepackage{booktabs}
\usepackage{bbm}
\usepackage{subcaption}
\usepackage{tabularx,ragged2e}
\usepackage{multicol,multirow}
\usepackage{enumitem}
\usepackage{soul}
\usepackage{arydshln}
\usepackage{bm}
\usepackage{pifont}

\definecolor{carminered}{rgb}{1.0, 0.0, 0.22}
\definecolor{coralred}{rgb}{0.93, 0, 0}

\NewTColorBox{NewBox}{ s O{!htbp} }{%
  floatplacement={#2},
  IfBooleanTF={#1}{float*,width=\textwidth}{float},
  }

\usepackage{microtype}

\title{Accelerating LLaMA Inference by Enabling Intermediate Layer Decoding via Instruction Tuning with LITE}

\author{Neeraj Varshney \hspace{14pt} Agneet Chatterjee \hspace{14pt} Mihir Parmar \hspace{14pt} Chitta Baral
  \\
  Arizona State University 
  }

\begin{document}
\maketitle
\begin{abstract}
Large Language Models (LLMs) have achieved remarkable performance across a wide variety of natural language tasks; however, their large size makes their inference slow and computationally expensive. 
Focusing on this problem, we propose to 
instruction tune LLMs with additional explicit \textbf{L}osses from the \textbf{I}n\textbf{T}ermediate lay\textbf{E}rs (\texttt{LITE}) and show that it enables these layers to acquire `good' generation ability without affecting the generation ability of the final layer.
We perform `\textit{dynamic confidence-based early exiting}' at {token level} from the intermediate layers which improves the efficiency of text generation without compromising the quality of the generation.
We conduct comprehensive experiments by instruction tuning LLaMA-2 models on the Alpaca dataset and holistically evaluate on four different human-instruction test sets.
We show that \textbf{dynamic early exiting achieves consistent and considerable inference computation cost improvements (\bm{$37.86\%$} for 7B and \bm{$46.35\%$} for 13B model) while maintaining the generation quality of the responses}.
We further conduct a thorough analysis of the results over several important aspects, such as comparing the semantic similarity of the outputs and dissecting the efficiency improvements by comparing the number of tokens generated in the output.
In summary, our work contributes to improving the efficiency of LLM inference while maintaining the generation quality, a crucial step en route to enabling their widespread adoption.

\end{abstract}

\section{Introduction}
\label{sec_introduction}
Recently developed Large Language Models (LLMs) \cite{touvron2023llama,NEURIPS2020_1457c0d6,chowdhery2022palm,rae2021scaling,smith2022using} have revolutionized the field of Natural Language Processing and achieved remarkable performance across a wide variety of tasks ranging from text generation and question answering to code generation and complex reasoning.
`Instruction Tuning' further teaches these language models to follow the user's instruction provided in natural language \cite{wei2022finetuned,mishra-etal-2022-cross,sanh2022multitask,wang-etal-2022-super,chung2022scaling}.
Despite all the notable abilities of these models, their large size (number of parameters) makes their inference
slow and computationally expensive which 
poses a practical challenge limiting their widespread adoption in resource constrained real-world applications.
Focusing on the above problem, in this work, 
\textbf{we propose to instruction tune LLMs in a way that enables intermediate layer decoding
for efficiently generating text without compromising the quality of the generation.}

We first show that in standard instruction tuning, only the final layer of the model acquires the ability to generate `\textit{quality}' text while the representations of the intermediate layers (when passed through the language modeling head) fail to do so. 
This restricts decoding from these intermediate layers
without degrading the generation quality.
Addressing this point, we propose to instruction tune LLMs with additional explicit \textbf{L}osses from the \textbf{I}n\textbf{T}ermediate lay\textbf{E}rs (\texttt{LITE}) and show that it enables these layers to acquire `good' generation ability. 
Importantly, we show that these layers acquire this ability without affecting the generation ability of the final layer; however, as expected, their generation ability still remains slightly inferior to the generation ability of the final layer. 
Thus, decoding the \underline{complete response} from intermediate layers improves the efficiency of inference but still results in minor degradation in the quality of the response.

\begin{figure*}[htbp]
\centering
    \begin{subfigure}{0.7\linewidth}
        \includegraphics[width=\linewidth]{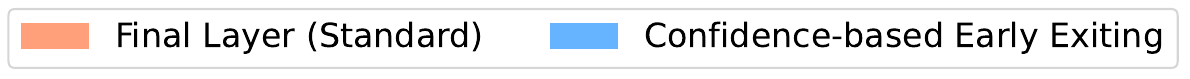}
    \end{subfigure}

    \begin{subfigure}{.24\linewidth}
        \includegraphics[width=\linewidth]{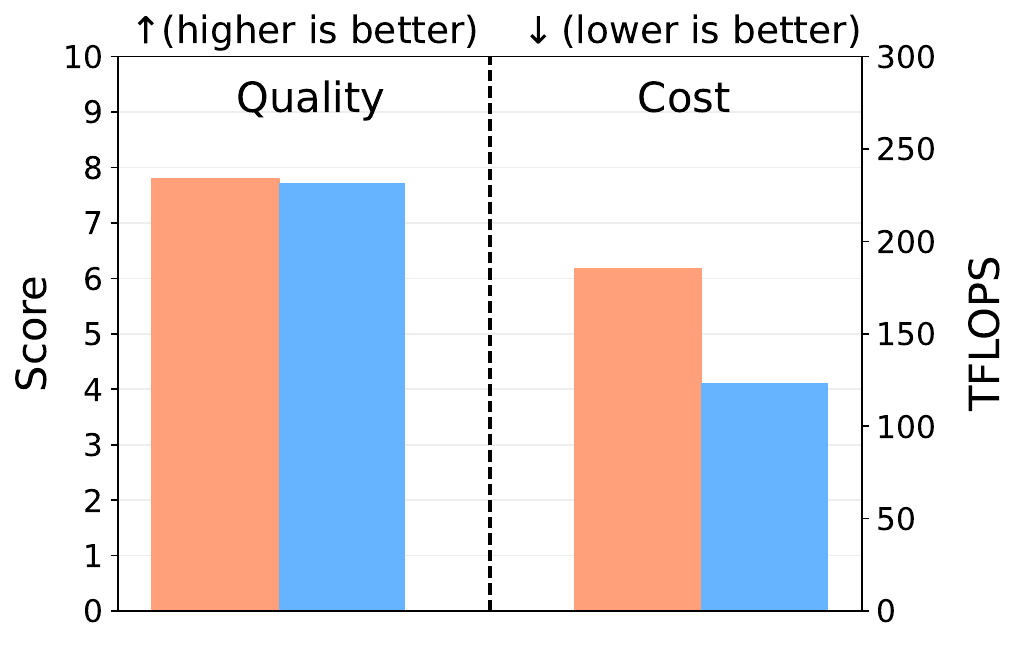}        
        \captionsetup{labelformat=empty}
        \caption{Vicuna (7B)}        
    \end{subfigure}
    \begin{subfigure}{.24\linewidth}
         \includegraphics[width=\linewidth]{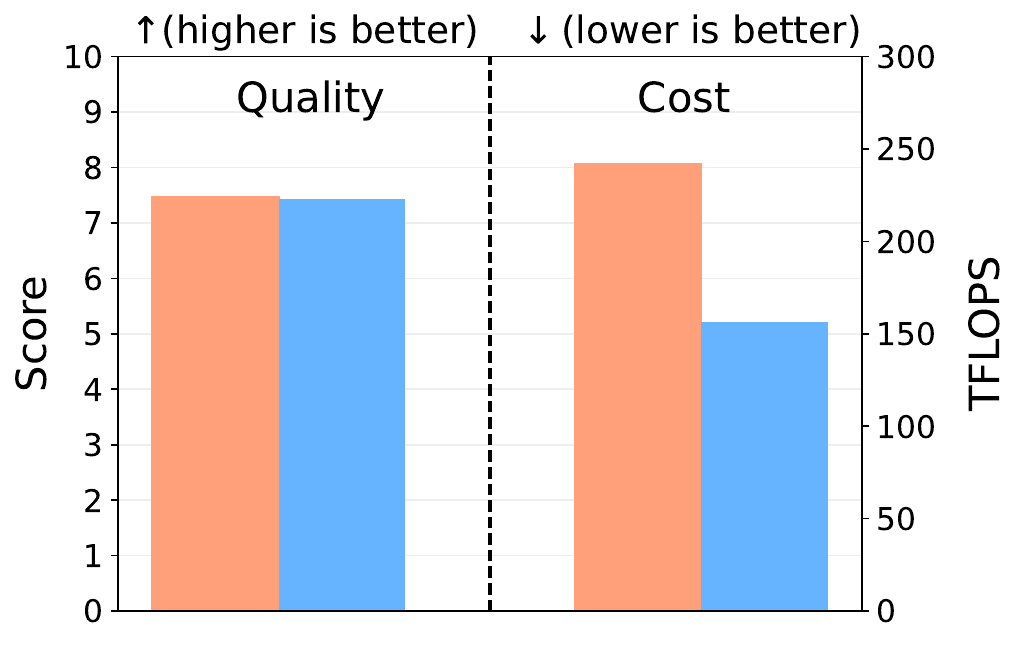}     
         \captionsetup{labelformat=empty}
         \caption{Koala (7B)}
    \end{subfigure}    
    \begin{subfigure}{.24\linewidth}        
         \includegraphics[width=\linewidth]{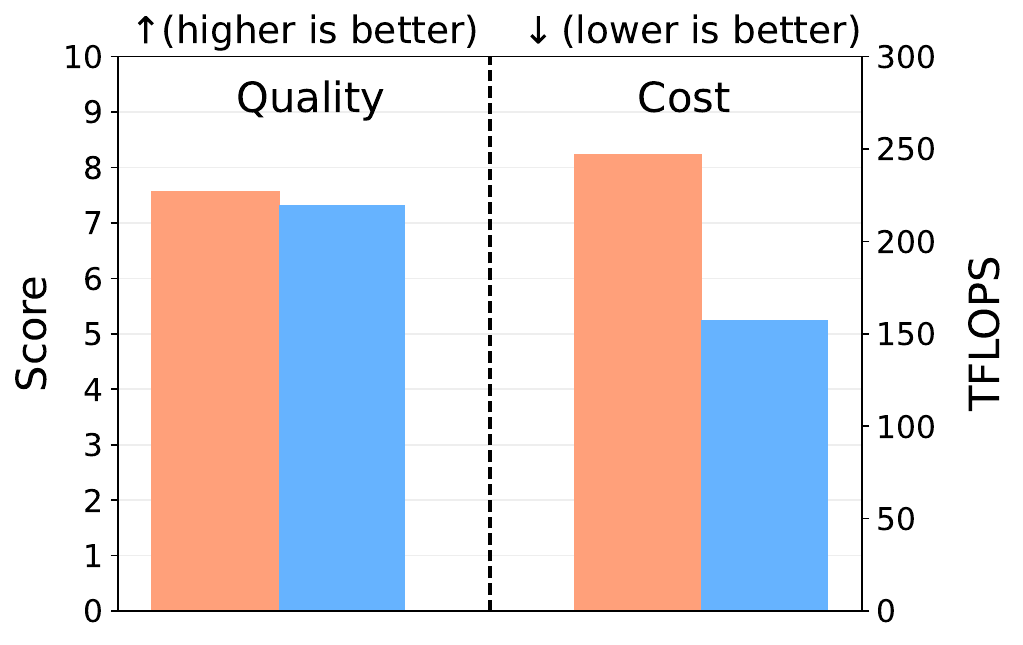}      
         \captionsetup{labelformat=empty}
         \caption{WizardLM (7B)}
    \end{subfigure}    
    \begin{subfigure}{.24\linewidth}
         \includegraphics[width=\linewidth]{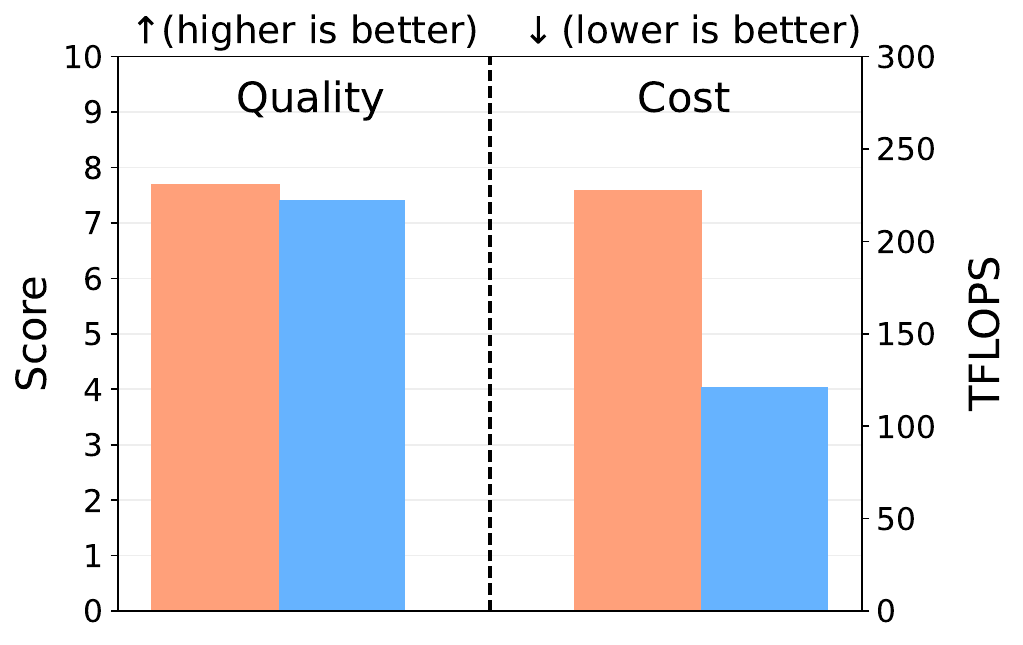}  
         \captionsetup{labelformat=empty}
         \caption{Self-Instruct (7B)}        
    \end{subfigure}

    \begin{subfigure}{.24\linewidth}
        \includegraphics[width=\linewidth]{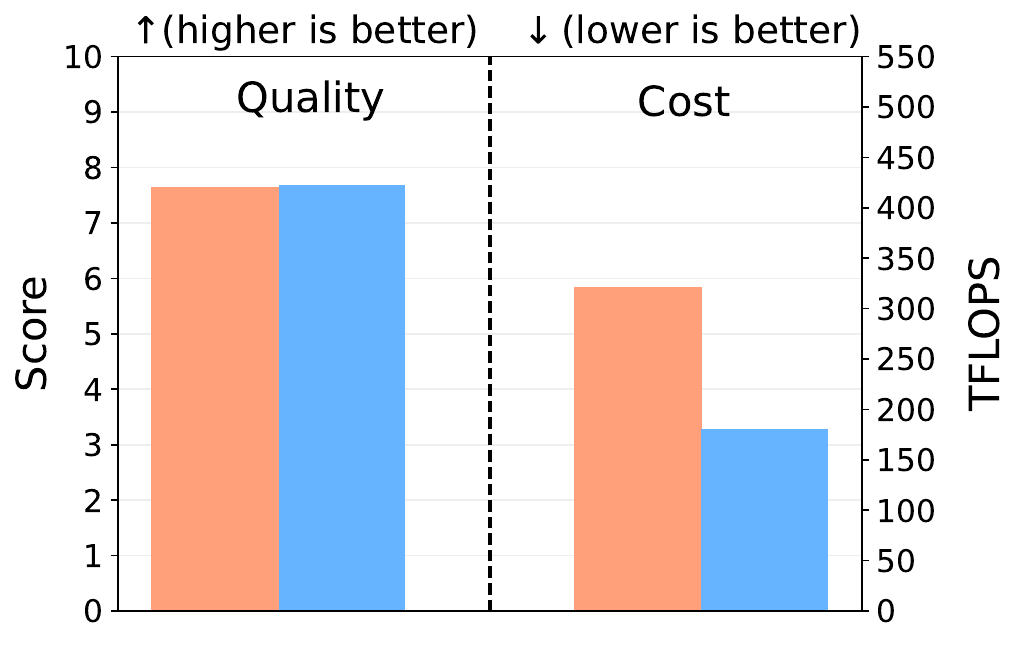}       
        \captionsetup{labelformat=empty}
        \caption{Vicuna (13B)}        
    \end{subfigure}
    \begin{subfigure}{.24\linewidth}
         \includegraphics[width=\linewidth]{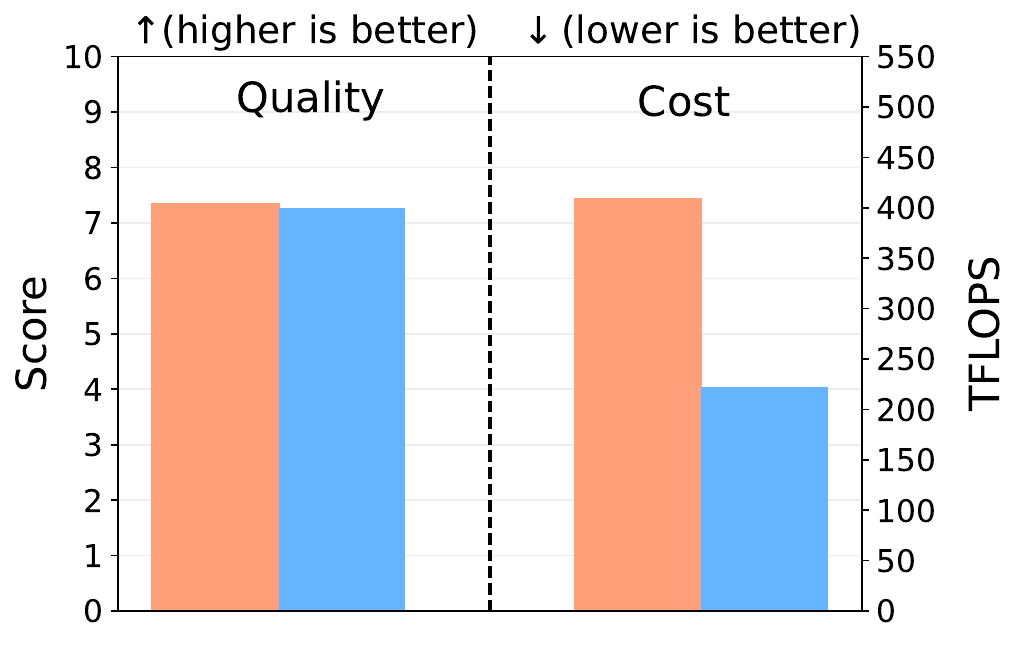}  
         \captionsetup{labelformat=empty}
         \caption{Koala (13B)}
    \end{subfigure}    
    \begin{subfigure}{.24\linewidth}        
         \includegraphics[width=\linewidth]{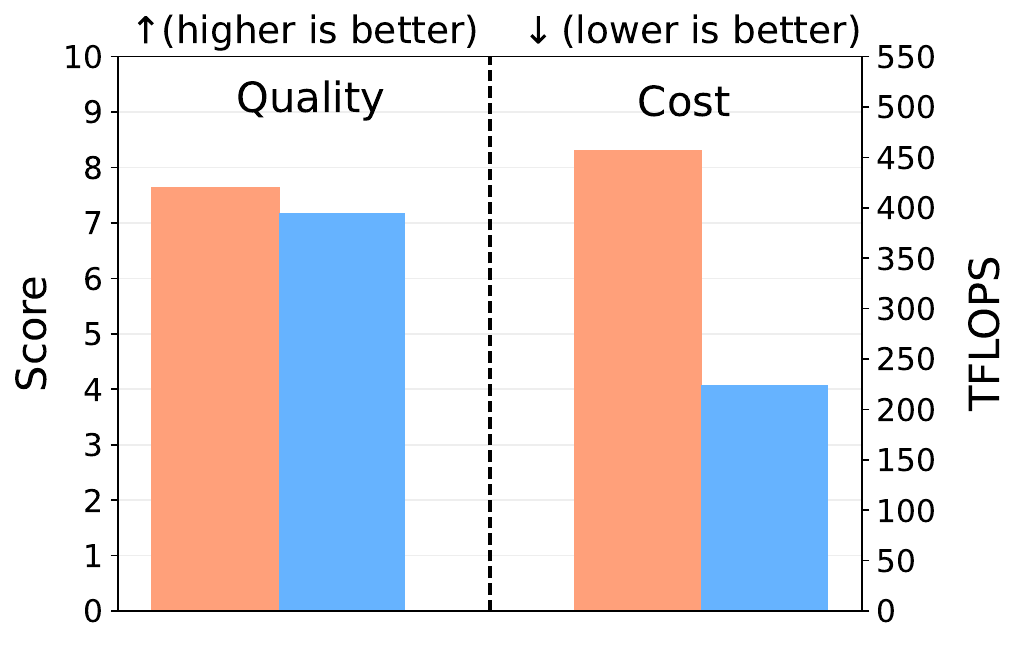}      
         \captionsetup{labelformat=empty}
         \caption{WizardLM (13B)}
    \end{subfigure}    
    \begin{subfigure}{.24\linewidth}
         \includegraphics[width=\linewidth]{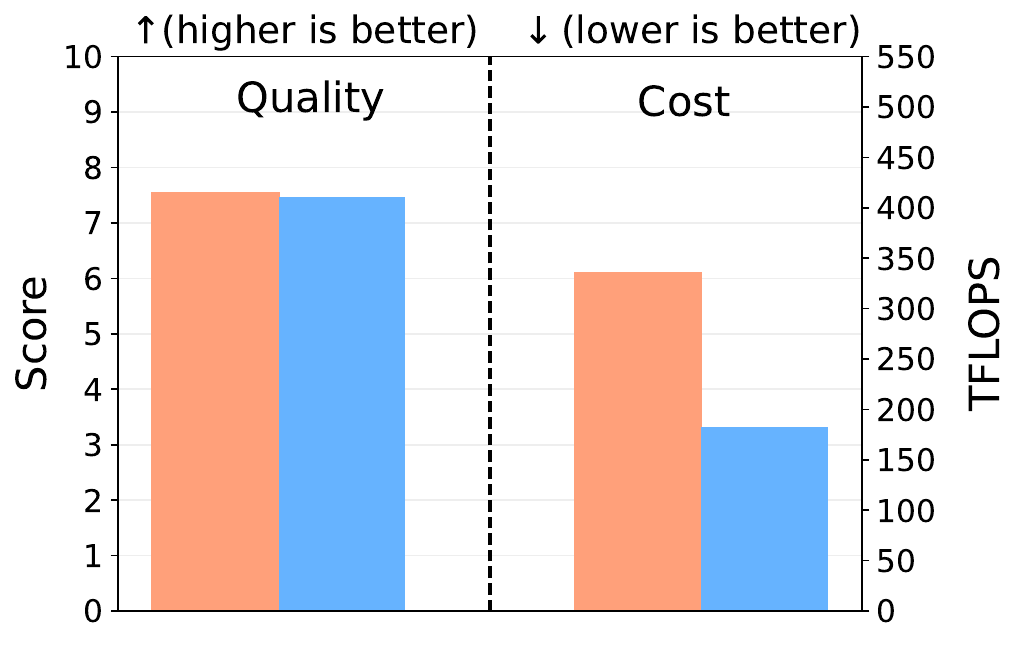} 
         \captionsetup{labelformat=empty}
         \caption{Self-Instruct (13B)}        
    \end{subfigure}
    
    \caption{Comparing the quality of responses (evaluated using the Claude model) and the inference cost (measured in FLOPs) of the standard generation from the final layer with our dynamic early exiting method. 
    It shows that {the dynamic early exiting achieves consistent and considerable cost improvements ({$37.86\%$} for 7B and {$46.35\%$} for 13B model on average) without degrading the generation quality}.
    For this study, the LLaMA-2 models (7B (top) and 13B (bottom)) are instruction tuned on the Alpaca dataset and evaluated on four test sets.
    }
    \label{fig:teaser}    
\end{figure*}
Addressing the above limitation, we show that (a) \texttt{LITE} greatly aligns the intermediate layers' token prediction with that of the final layer and (b) the intermediate layers' token prediction probabilities provide a strong signal of this alignment.
Building on top of these two findings, \textbf{we perform `\textit{dynamic confidence-based early exiting}' at \underline{token level} from the intermediate layers which improves the efficiency of inference while maintaining the generation quality}.

We conduct comprehensive experiments by instruction tuning LLaMA-2 models \cite{touvron2023llama} on the widely used Alpaca dataset \cite{alpaca} and holistically evaluate on four different human-instruction test sets including Vicuna \cite{vicuna2023}, WizardLM \cite{xu2023wizardlm}, Koala \cite{koala_blogpost_2023}, and Self-Instruct \cite{wang2022self}.
Figure \ref{fig:teaser} compares the \textbf{quality of responses} (evaluated using the Claude model as detailed in Section \ref{sec_experimental_setup}) and the \textbf{inference cost} (measured in FLOPs) of the (i) standard generation method from the final layer with (ii) our dynamic early exiting method.
It shows that \textbf{dynamic early exiting achieves consistent and considerable inference cost improvements (\bm{$37.86\%$} for 7B and \bm{$46.35\%$} for 13B model on average) while maintaining the generation quality}.

We further perform a thorough analysis of the results by 
(a) studying the quality and inference cost comparison at a category level on the evaluation datasets (\ref{sec_category_level_analysis}),
(b) analyzing the semantic similarity between the responses generated from the final layer and the early exiting method (\ref{sec_semantic_similarity}), and
(c) dissecting the efficiency improvements of the dynamic early exiting method 
by comparing the number of tokens generated in the outputs (\ref{sec_number_of_tokens}).

In summary, we show that instruction tuning with additional explicit losses from the intermediate layers (\texttt{LITE}) enables the intermediate layers to acquire `good' generation ability and our `dynamic early exiting' method leverages that to improve the efficiency of inference while maintaining the generation quality.
We further discuss the potential of intermediate layer decoding in `\textit{speculative sampling}' and `\textit{hallucination detection}'.

\section{Related Work}
\label{sec_related_work}
Improving the inference efficiency of large language models is an important research direction and is receiving considerable attention from the NLP community. 
In this section, we first review some of the existing methods and then detail their differences from our work.

\textbf{Reducing model size:} Since model size plays a crucial role in increasing the inference cost and latency, techniques like \textbf{quantization} \cite{dettmers2022llm,yao2022zeroquant,pmlr-v202-xiao23c,frantar2022gptq}, \textbf{knowledge distillation} \cite{hsieh2023distilling,jiao-etal-2020-tinybert,li-etal-2022-dq,mirzadeh2020improved}, \textbf{model compression and network pruning} \cite{wang-etal-2020-structured,guo-etal-2021-parameter} have been shown to be effective in improving the inference efficiency.

Furthermore, during sampling, a cache of the keys and values can be maintained for every attention layer which reduces the computations at inference time (\textbf{KV caching}). 
However, it increases the GPU VRAM memory requirement of inference.

Another technique to improve inference efficiency is \textbf{speculative sampling} \cite{leviathan2023fast, chen2023accelerating} in which a short draft of $K$ tokens is first generated from a smaller (thus faster) auto-regressive model. Then, the draft is scored using the larger model which corresponds to the target model from which we wish to sample from.
Using some rejection sampling scheme, a subset of the $K$ draft tokens is accepted by sequentially checking from left to right and thus in this process, we recover the distribution of the target model for the accepted tokens.
The efficiency in this technique comes from `producing' more than one token (on average) from the target model in a single pass.

\textbf{Early exiting and cascading} based inference techniques have been shown to be effective for classification tasks with BERT-style models, 
such as DeeBERT \cite{xin-etal-2020-deebert} that speeds up BERT inference by inserting extra classification layers between each encoder layer, PoWER-BERT \cite{goyal2020power} that focuses on progressive word-vector elimination (based on significance computed using self-attention) along the encoder pipeline, DynaBERT \cite{NEURIPS2020_6f5216f8} that adjusts the size of the model by selecting adaptive width and depth, and cascading \cite{varshney-baral-2022-model, li-etal-2021-cascadebert-accelerating,varshney-baral-2023-post,yue2023large,cheng2023batch} in which sequential inference is done through models of bigger and bigger size with conditional exiting to output predictions in an efficient yet accurate manner.
Our work is also related to Confident Adaptive Language Modeling (CALM) \cite{schuster2022confident} and Depth-Adaptive Transformers \cite{Elbayad2020Depth-Adaptive} in which early exiting is performed by learning additional classifiers attached to the decoder layers.  

\textbf{Casting internal representations:} 
\citet{din2023jump} proposed to short-cut away transformer inference in between certain layers by learning linear transformations across layers in the network.

\textbf{Contrastive decoding from intermediate layers:} 
Prior work has also explored leveraging the intermediate layers for contrastive decoding to improve reasoning \cite{o2023contrastive,gera2023benefits}.

Our work differs from existing work in
the following aspects: 
\begin{enumerate}[noitemsep,nosep,leftmargin=*]

    \item Most of the existing early exiting works focus on improving the efficiency of encoder-only models (like BERT) or encoder-decoder models (like T5); \textbf{our work focuses on the current state-of-the-art decoder-only LLMs (LLaMA-2).
    Furthermore, we focus on the instruction tuning setting}, unlike prior work that focused on solving specific tasks like GLUE classification tasks, summarization, or QA. 
    We note that the open-ended generation is a much more challenging setting for intermediate layer exiting than classification or MCQ tasks.

    \item Early exiting methods typically require training additional classifiers for the intermediate layers, however, \textbf{in our method, we use the same shared language modeling head at all the layers, thus, do not introduce new model parameters}.

    \item \textbf{For leveraging the intermediate layers for decoding, we enable them to acquire good generation ability by instruction tuning with LITE}, unlike other methods that use a pre fine-tuned model in which these layers have poor generation ability.

    \item Most of the existing methods typically require complex architectural modifications, network pruning, saliency quantification, knowledge distillation, or adding extra model parameters. In contrast, \textbf{our proposed method (both for instruction tuning and inference) is simple and easy to implement as it requires just a few modifications during tuning and inference} and yet it delivers considerable improvements. 
    Furthermore, our method is complementary to these existing methods.

    \item The computational efficiency in a lot of existing methods often comes with a compromise in performance. Contrary to this, we show that \textbf{our dynamic early exiting method maintains the generation quality while providing efficiency benefits}.

    \item Some existing methods typically require training a separate model for each computation budget; on the other hand, \textbf{in our method, the same model can be adjusted to meet all the computation constraints (by varying the confidence thresholds for exiting)}.

\end{enumerate}

\section{Instruction Tuning with Losses from Intermediate Layers}

In this section, we first detail the standard instruction tuning (\texttt{IT}) and then describe instruction tuning with additional explicit losses from the intermediate layers (\texttt{IT} with \texttt{LITE}).

\begin{figure}[t!]
    \centering
    \includegraphics[width=7.5cm]{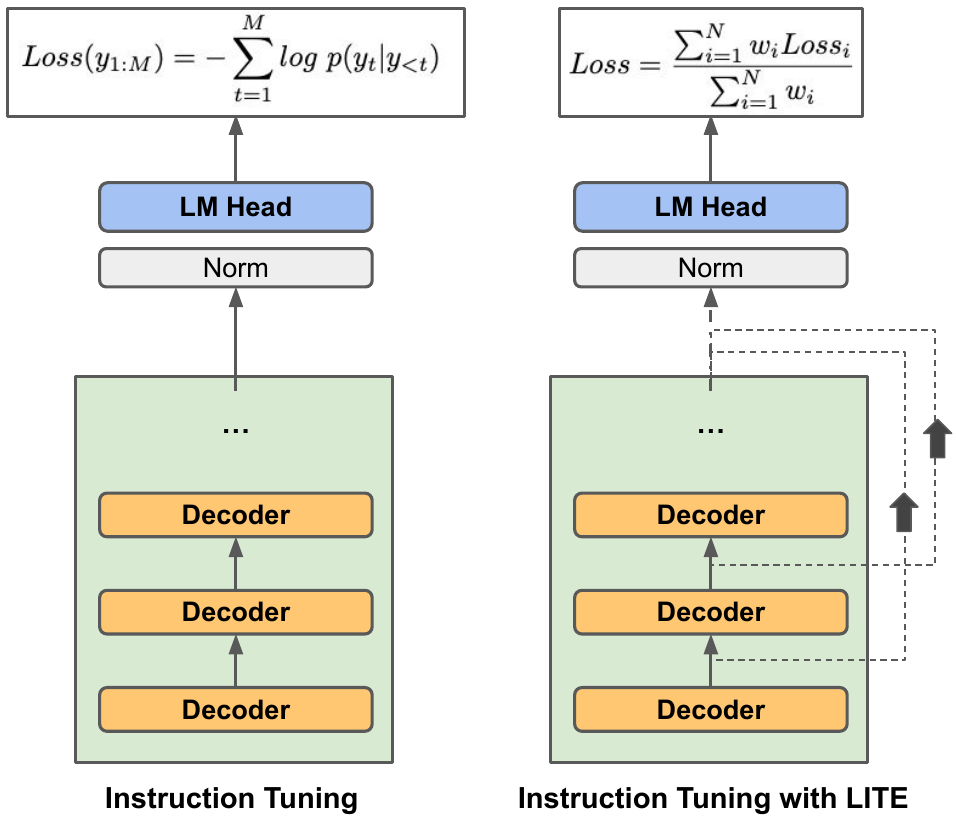}
    \caption{Loss calculation for standard instruction tuning (left) and instruction tuning with additional explicit losses from the intermediate layers \texttt{LITE} (right).}
    
    \label{fig:model_architecture}
\end{figure}

\paragraph{Instruction Tuning (\texttt{IT}):}
One of the major reasons that necessitate instruction tuning of LLMs is the 
mismatch between their pre-training objective and the users’ objective, i.e., LLMs are typically trained on minimizing the word prediction error on large corpora; while users want the model to follow their instructions.
To this end, an instruction tuning dataset typically consisting of (instruction, input, and output) triplets is collected and a pre-trained model is fine-tuned in a fully supervised manner, where given the instruction and the input, the model is trained by predicting each token in the output \cite{mishra-etal-2022-cross,chung2022scaling, wei2021finetuned,wang-etal-2022-super,parmar-etal-2022-boxbart}.
Loss calculation during instruction tuning of a typical decoder-only LLM (LLaMA in this case) is shown in Figure \ref{fig:model_architecture} (left).
The model consists of a stack of decoder layers followed by a language modeling head which outputs the probability distribution over the vocabulary tokens as its prediction.
During the supervised fine-tuning, the loss over the output tokens is backpropagated from the final layer of the model:
\[
     Loss(y_{1:M}) =  - \sum_{t=1}^{M} log \hspace{3pt}p(y_t | y_{<t})
\]

\paragraph{Instruction Tuning with \underline{L}osses from \underline{I}n\underline{t}ermediate Lay\underline{e}rs (\texttt{IT} with \texttt{LITE}):}
We show that in the standard instruction tuning, only the final layer of the model acquires the ability to generate `quality' text while the representations of the intermediate layers (when passed through the language modeling head) fail to do so (Section \ref{sec_inability_of_regular_LLM}). 
In other words, it does not explicitly teach the intermediate layers of the tuned LLM to generate tokens. This restricts decoding from these intermediate layers without degrading the generation quality.

We note that during tuning, the same language modeling head (that is used with the final layer) can also be used with the intermediate layers to obtain the losses of those layers. 
Note that this does not impact the number of parameters of the model as the same shared language modeling head is used for obtaining these losses.
To this end, we calculate a weighted aggregation of the losses from the intermediate layers (including the final) to calculate the overall loss value as shown below:

\[
     Loss = \frac{ \sum_{i=1}^{N} w_{i} {Loss}_i }{\sum_{i=1}^{N} w_{i}}
\]

where $N$ is the number of layers, $w_i$ is the weight corresponding to the $i^{th}$ layer, and ${Loss}_i$ is the cross entropy loss of the $i^{th}$ layer as shown in Figure \ref{fig:model_architecture}.

During tuning, we use the representations of the intermediate layers and calculate the loss from these layers at the end.  
We note that this is a general formulation as it captures a variety of scenarios including the standard fine-tuning in which the loss is calculated only from the last layer (i.e., $w_i$ = 0 for $i$=1 to $N$-1 and $w_N$ = 1).
Furthermore, this formulation also allows aggregating losses from only the selected intermediate layers instead of all the layers by accordingly defining the LM head pathways and the $w_i$ values.
In Section \ref{sec_LITE_has_no_adverse_effect}, we will show that \texttt{IT} with \texttt{LITE} while enabling the intermediate layers with `good' generation ability does not adversely affect the final layer's generation ability.
Furthermore, as expected, the quality of generation typically improves with the layer number as the later layers have more capacity to learn.

\section{Making Inference Efficient}

In this section, we first detail auto-regressive inference and then describe early exiting techniques, namely, fixed early exiting (Section \ref{sec_fixed_exiting}) and dynamic confidence-based early exiting (Section \ref{sec_dynamic_exiting}).

\paragraph{Auto-Regressive Inference:}
In the context of language models, auto-regressive inference refers to the process of generating sequence of tokens where each token is generated based on the preceding tokens in the sequence.
For generating a token, the model takes the input (including the previously generated tokens) and runs a forward pass in which the input is fed to the model and passed sequentially along its layers until the probabilities for the next token are predicted (called as \textbf{logits}).
Chaining model forward passes with next token selection iteratively leads to the generation of text.

Auto-regressive language generation is based on the assumption that the probability distribution of a token sequence can be decomposed into the product of conditional next token distributions:
\[
     P(w_{1:T} | W_0) = \prod_{t=1}^{T} P(w_t | w_{1:t-1}, W_0)
\]

where $W_0$ corresponds to the given input prompt and the length $T$ is usually determined on the fly when the token \texttt{EOS} is generated or when the specified maximum number of tokens have been generated.
In \textbf{greedy decoding}, the token with the highest probability is selected as the next word prediction at each timestep t.
\[
     w_t = \texttt{argmax}_w P(w | w_{1:t-1}, W_0)     
\]

During sampling, a cache of the keys and values can be maintained for every attention layer (called {KV caching}) which reduces the computations required at inference time. 
However, this increases the GPU VRAM memory requirement of inference.

In the following subsections, we describe early exiting techniques for efficient inference.

\subsection{Fixed Early Exiting}
\label{sec_fixed_exiting}

Since instruction tuning with \texttt{LITE} enables the intermediate layers to acquire `good' generation ability,
the computations during inference can be terminated at a pre-specified intermediate layer (referred to as \textbf{exiting layer}) and the language modeling head can be used to predict the next token.
This saves the computations of the remaining layers that follow the specified exiting layer and thus it improves the efficiency of inference. 

Though this method of fixed early exiting leads to improvement in the efficiency of inference, it is bound to result in some degradation in the quality of the generation as the generation ability of an intermediate layer still remains inferior to the generation ability of the final layer.
We also note that the quality of generation typically improves with the layer number as the later layers have more capacity and hence ability.

\subsection{Dynamic Confidence-Based Early Exiting}
\label{sec_dynamic_exiting}

Addressing the limitation of the fixed early exiting method, we study a dynamic early exiting method that decides the exiting layer for a token prediction based on the intermediate layer's probability of the prediction (softmax over the logit values).

This is motivated by our following two findings:

(a) Instruction Tuning with \texttt{LITE} greatly aligns the intermediate layers' token prediction with that of the final layer (Section \ref{sec_LITE_aligns_tokens}) and 

(b) The intermediate layers' token prediction probabilities (referred to as \textbf{confidence}) provide a strong signal of this alignment (Section \ref{sec_confidence_alignment}).

Building on top of these two findings, we perform `\textit{dynamic confidence-based early exiting}' at \underline{token level} from the intermediate layers which improves the efficiency of inference while maintaining the generation quality.
Specifically, 
a set of intermediate layers with their corresponding confidence thresholds are defined and at inference time, the exiting decision for a prediction is taken by comparing the intermediate layer's prediction confidence against its corresponding threshold.
This enables the model to do efficient inference without degrading the quality of generation.

Some prior early exiting methods learn new classifiers for the intermediate layers to make the exiting decision. Here, we do not introduce new parameters and use the softmax probability to make the exiting decision.

Note that we explore this exiting method for inference without using KV caching.
This is because the standard KV caching cannot be used here as the representations of the layers after the exiting layer are not computed in this method and thus will not be available in the cache for the next token prediction if the model exits from a higher layer than the previous token prediction.

\section{Experimental Setup}
\label{sec_experimental_setup}

\paragraph{Tuning and Inference:}
We instruction tune the LLaMA-2 models \cite{touvron2023llama} (7B and 13B) with the widely used Alpaca dataset \cite{alpaca}.
Alpaca consists of 52K instruction-following demonstrations generated from OpenAI’s \texttt{text-davinci-003} using the self-instruct \cite{wang2022self} technique.
In IT with \texttt{LITE} for 7B model (32 total layers), we aggregate losses from the following selected intermediate layers: (8, 12, 16, 20, 24, 28) along with the final layer and use equal weights in loss calculation.
Similarly, for the 13B model (40 total layers), we use (8, 12, 16, 20, 24, 28, 32, 36) layers.
We do full parameter fine-tuning on 4 A100 GPUs.

We skip selecting the initial layers because they have a limited capacity to learn and do not result in good token predictions.
Furthermore, we select layers at an interval of 4 so that at inference time, the model can do enough reasoning/interactions between two consecutive checkpoints. 
Otherwise, checking at all every layer can result in computational overhead.
We train this model for 5 epochs so that it achieves training loss comparable to standard tuning.
For inference, we set the max new tokens value to 256. 

We present all the results corresponding to this tuning configuration and leave the exploration of selecting different intermediate layers and different weights for these layers for future work.

\paragraph{Evaluation Datasets:}
\begin{table}[t]
    \centering
    {
    \begin{tabular}{@{}cc@{}}
        \toprule
         \textbf{Test Set} & \textbf{\# Samples}\\
         
        \midrule
        
        Vicuna & 80 \\
        Koala & 180 \\
        WizardLM & 218 \\
        Self Instruct & 252 \\
        
    \bottomrule
    \end{tabular}    
    }
    \caption{
    Statistics of evaluation datasets.
    }

    \label{tab:evaluation_test_sets}
\end{table}
To perform holistic evaluation, we experiment with four different human-instruction test sets including Vicuna \cite{vicuna2023}, Self-Instruct \cite{wang2022self}, Koala \cite{koala_blogpost_2023}, and WizardLM \cite{xu2023wizardlm}.
We select these evaluation test sets as they can together cover a large number and types of instructions thus resulting in a comprehensive evaluation.
Table \ref{tab:evaluation_test_sets} shows the statistics of the datasets.

\begin{figure*}[htbp]
\centering

    \begin{subfigure}{.24\linewidth}
        \includegraphics[width=\linewidth]{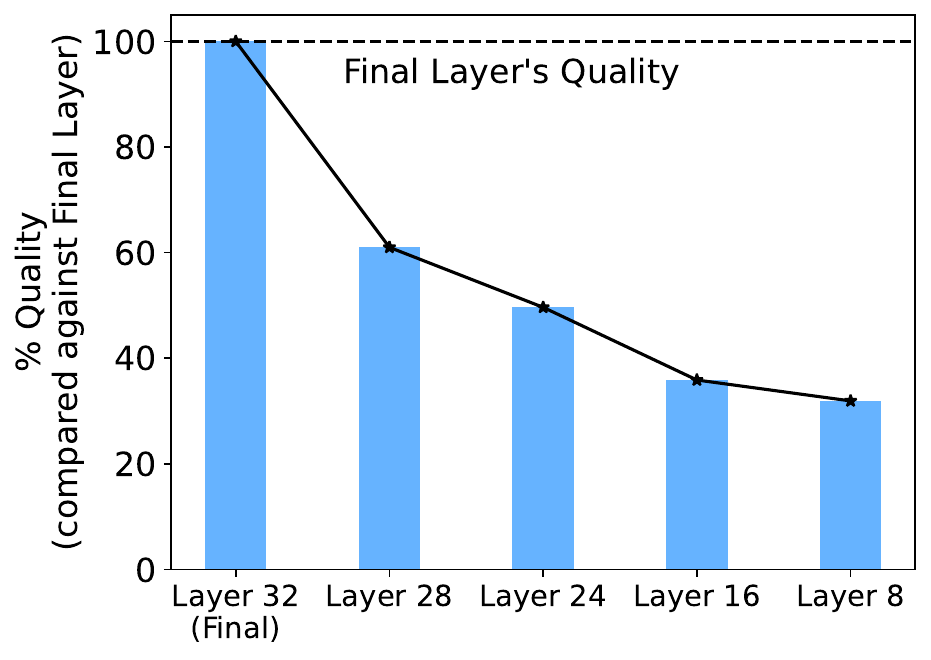}        
        \caption{Vicuna}        
    \end{subfigure}
    \begin{subfigure}{.24\linewidth}
         \includegraphics[width=\linewidth]{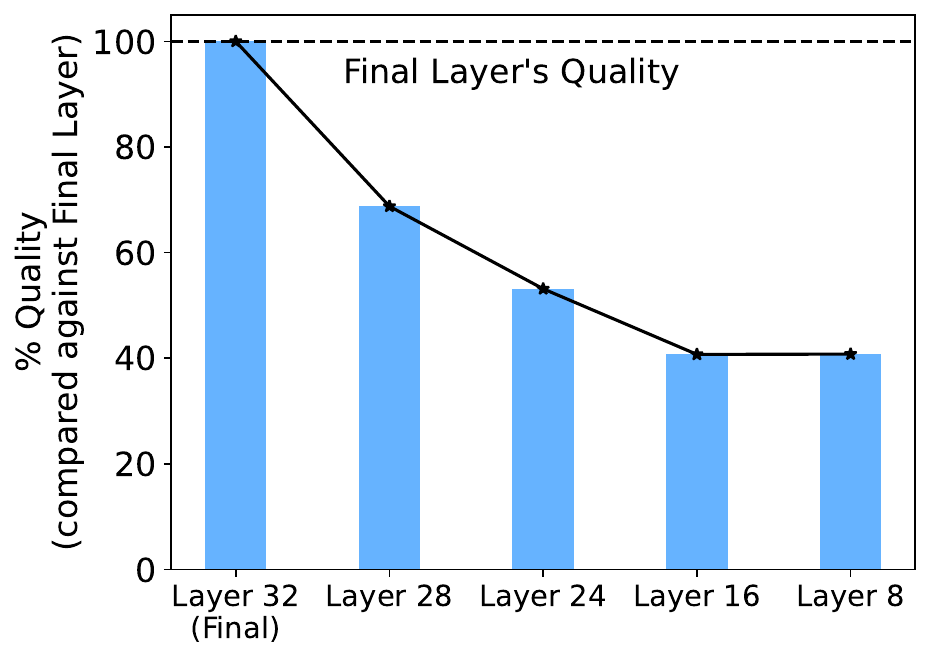}         
         \caption{Koala}
    \end{subfigure}
    \begin{subfigure}{.24\linewidth}        
         \includegraphics[width=\linewidth]{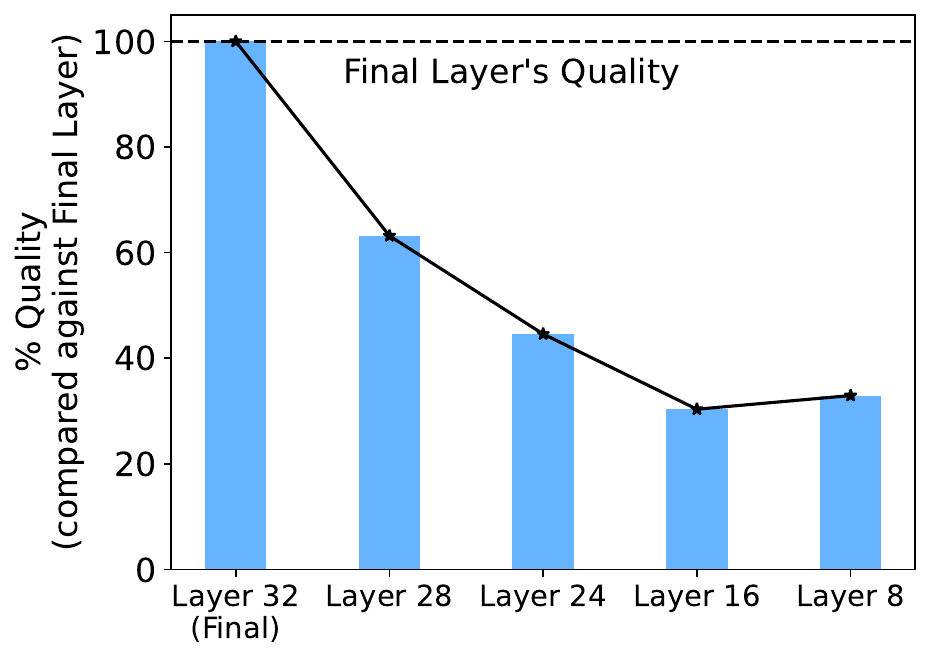}         
         \caption{WizardLM}
    \end{subfigure}    
    \begin{subfigure}{.24\linewidth}
         \includegraphics[width=\linewidth]{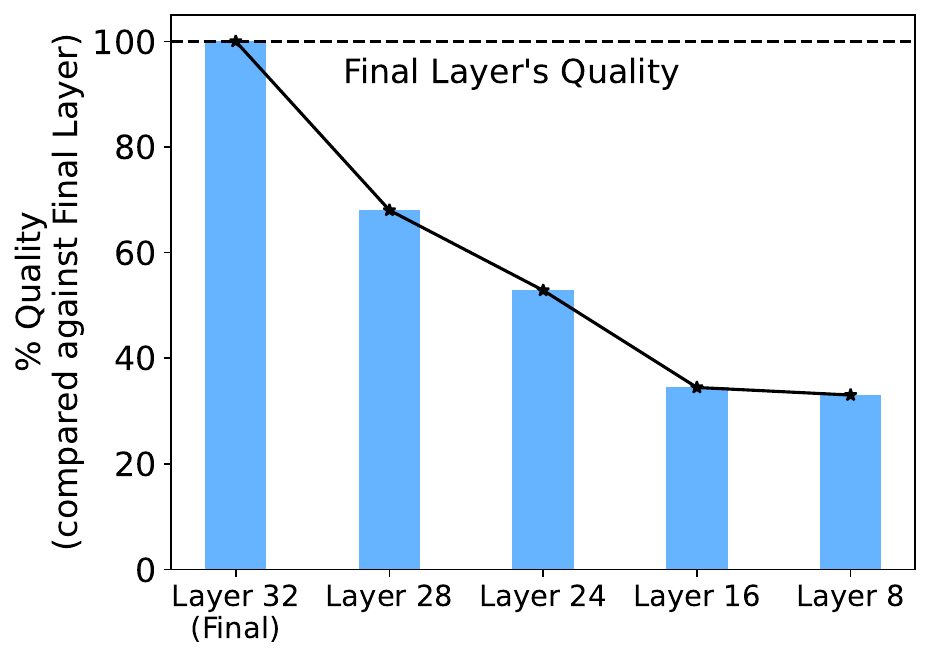}         
         \caption{Self-Instruct}        
    \end{subfigure}    
    \caption{Demonstrating quality comparison of the output of intermediate layers (generated via fixed exiting) against the final layer's generation of the model tuned with standard instruction tuning. 
    It shows that \textbf{the intermediate layers generate text of considerably degraded quality and this quality drops as the layer number decreases.}}
    \label{fig:quality_comparison_layer_wise_regular_model}    
\end{figure*}

\paragraph{Evaluation Methodology:}

The evaluation of the instruction-following ability of LLMs is challenging due to the existence of multiple correct responses to an input and the infeasibility of reproducing human evaluations. 
Addressing this problem, recent works have started to rely on automatic evaluations using LLMs \cite{zheng2023judging,vicuna2023}.
Specifically, an LLM like GPT-4 \cite{OpenAI2023GPT4TR} or Claude \cite{bai2022constitutional} is used as a judge to compare the quality of responses of two models on a given instruction.

We note that these LLMs have been shown to be vulnerable to position bias in their judgment \cite{wang2023large}. To circumvent this bias, we evaluate a response pair with both orderings of the responses and then aggregate the judgment scores.
We provide the prompt for comparing the quality of the responses of two models in Appendix \ref{evaluation_prompt}.

\section{Results}

In this section, we first demonstrate the inability of the intermediate layers of the model tuned with standard instruction tuning to generate `quality' text (\ref{sec_inability_of_regular_LLM}).
Then, we proceed to show the impact of instruction tuning with \texttt{LITE}.
Specifically, we first show that instruction tuning with \texttt{LITE} does not adversely affect the generation quality of the final layer (\ref{sec_LITE_has_no_adverse_effect}).
Then, 
we show that \texttt{LITE} aligns the intermediate layers' token predictions with the final layer (\ref{sec_LITE_aligns_tokens}), and the corresponding prediction confidence values provide a strong signal of this alignment (\ref{sec_confidence_alignment}).
These two findings motivate dynamic confidence-based early exiting method.
Finally, we show the effectiveness of early exiting in improving the efficiency of inference while maintaining the generation quality (\ref{sec_performance_early_exiting}).

To avoid repetition in the main paper, we present all the results corresponding to the 7B model variant unless otherwise mentioned. 
We present detailed results for the 13B model in Appendix \ref{sec_13b_results}.

\subsection{Inability of the Intermediate Layers of the Model Tuned with Standard Instruction Tuning to Generate `High-Quality' Text}
\label{sec_inability_of_regular_LLM}

In order to obtain the text (sequence of tokens) generated via fixed exiting from an intermediate layer, we apply the normalization (RMSNorm) followed by the language modeling head to the representations of that intermediate layer and skip the computations of the layers following the exiting layer (as detailed in Section \ref{sec_fixed_exiting}). 
For the model tuned with the standard instruction tuning, we compare the quality of the text generated from different intermediate layers against the final layer's generation in Figure \ref{fig:quality_comparison_layer_wise_regular_model}. 
We compare the quality using the Claude model as detailed in Section \ref{sec_experimental_setup}. 
As expected, the intermediate layers generate text of considerably degraded quality and this quality drops as the layer number decreases.

\textbf{This demonstrates that with standard instruction tuning, only the later layers (especially the final layer) of the model acquires the ability to generate `quality' text while the representations of the intermediate layers (when passed through the language modeling head) fail to do so. }

Thus, for the model tuned with standard instruction tuning, the early exiting inference method saves the inference computation cost but considerably degrades the quality of the generation.
This restricts employing such early exiting techniques for the model tuned with standard instruction tuning.
We show examples of responses obtained via fixed early exiting from different intermediate layers of the model in Appendix \ref{sec_appendix_inability_of_intermediate}.

We perform instruction tuning with \texttt{LITE} to enable the intermediate layers to acquire `good' generation ability. 
Importantly, we note that these layers acquire this ability without affecting the generation ability of the final layer as we will show in the next subsection.

\subsection{IT with LITE Does NOT Adversely Affect the Final Layer's Generation Ability}
\label{sec_LITE_has_no_adverse_effect}

\begin{figure}[htbp]
    \centering
    \includegraphics[width=6cm]{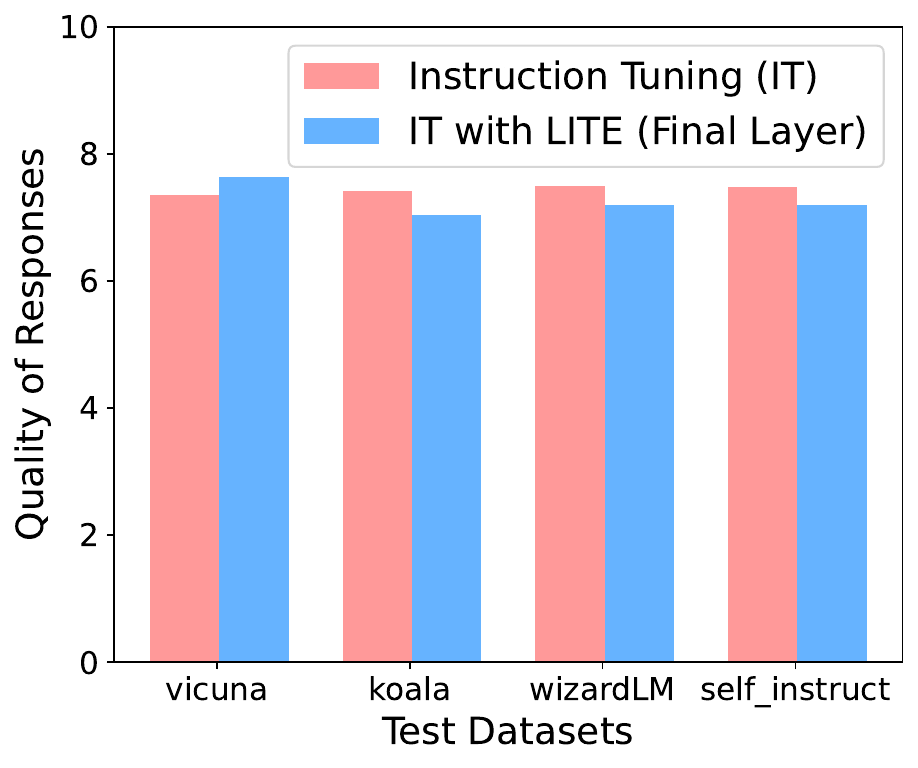}
    \caption{Comparing quality of responses (judged on a scale of 1 to 10 by Claude Model) of (a) the model tuned using IT and (b) the model tuned using IT with \texttt{LITE}. 
    The outputs of the models are of comparable quality which highlights that \textbf{instruction tuning with LITE does not adversely affect the generation quality of the final layer of the model}.}    
    \label{fig:no_adverse_impact}
\end{figure}

In Figure \ref{fig:no_adverse_impact}, we compare the quality of responses (judged on a scale of 1 to 10 by the Claude Model) of (a) the model tuned using standard instruction tuning (IT) and (b) the model tuned using IT with \texttt{LITE}.
Note that the responses for both these models correspond to their respective final layer's output.

From the figure, it can be observed that for all the datasets, \textbf{the outputs of both models are of comparable quality which shows that tuning with \texttt{LITE} does not adversely affect the generation ability of the final layer of the model}.

Next, we demonstrate two important characteristics of instruction tuning with \texttt{LITE} (in \ref{sec_LITE_aligns_tokens} and \ref{sec_confidence_alignment}) that motivate us to study dynamic confidence-based early exiting from the intermediate layers.

\subsection{IT with LITE Greatly `Aligns' Intermediate Layer Token Predictions with that of the Final Layer}
\label{sec_LITE_aligns_tokens}

We define percentage `alignment' of a layer as the measurement of how often the token predictions of that layer match with the token predictions of the final layer (given equal prefixes).
For this study, we do not do early exiting, instead we just use the representation of each intermediate layer and pass it through the LM head to obtain the corresponding token prediction of each layer. 
Note that for generating the next token, we follow the standard generation methodology and append the predicted token of the last layer to the input. 
Essentially, we obtain the token prediction of all the layers given the same prefix.

\begin{figure}[!]
\centering
    \begin{subfigure}{0.5\linewidth}
        \includegraphics[width=\linewidth]{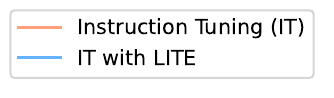}
    \end{subfigure}

    \begin{subfigure}{.48\linewidth}
        \includegraphics[width=\linewidth]{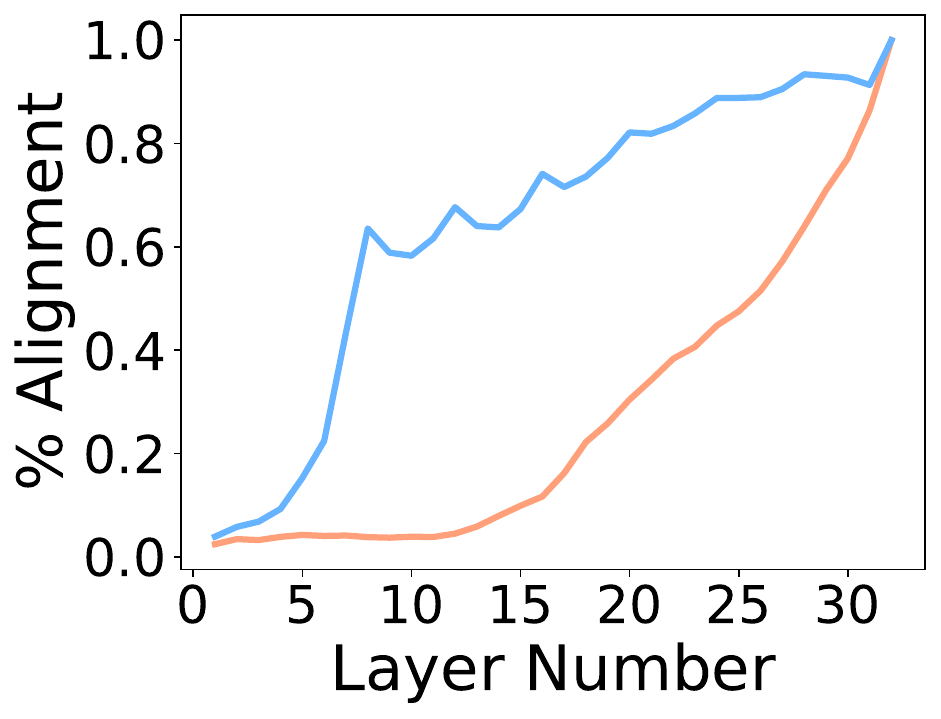}        
        \caption{Vicuna}        
    \end{subfigure}
    \begin{subfigure}{.48\linewidth}
         \includegraphics[width=\linewidth]{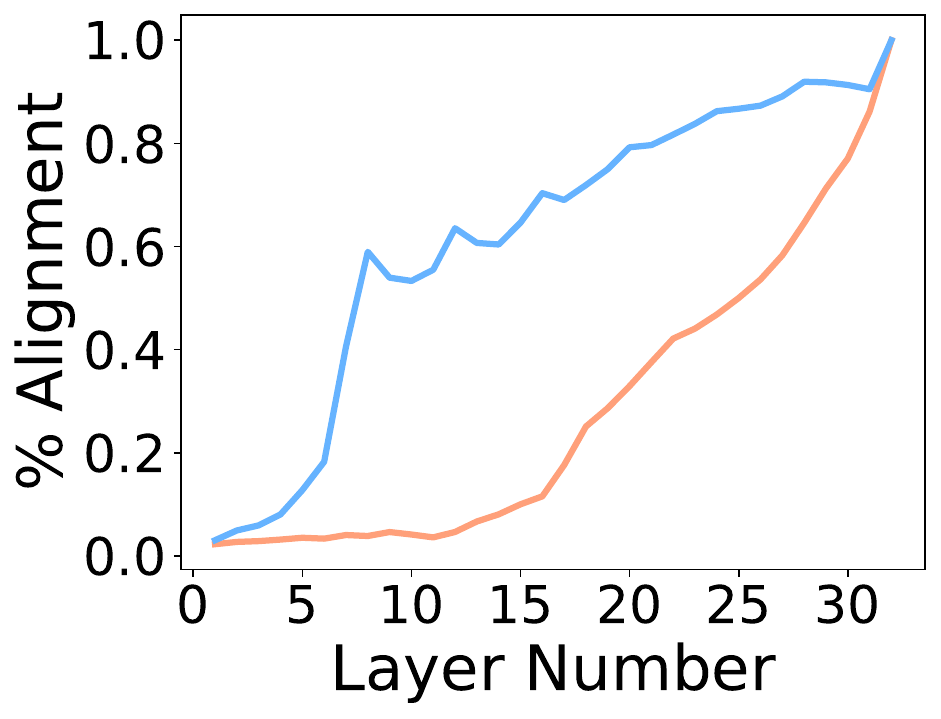}         
         \caption{Koala}
    \end{subfigure}
    
    \begin{subfigure}{.48\linewidth}        
         \includegraphics[width=\linewidth]{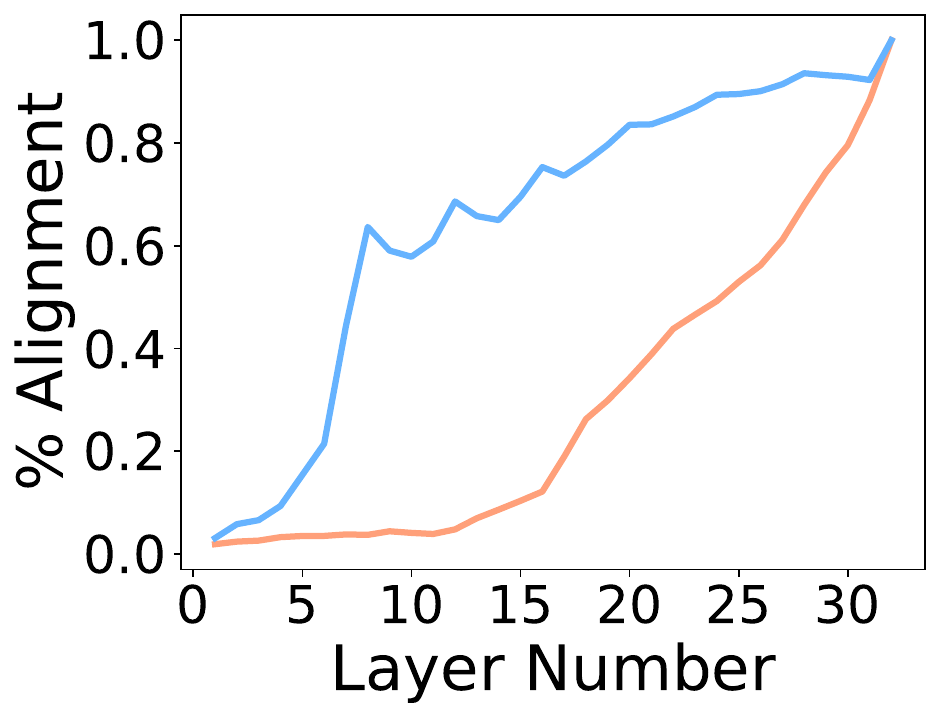}         
         \caption{WizardLM}
    \end{subfigure}    
    \begin{subfigure}{.48\linewidth}
         \includegraphics[width=\linewidth]{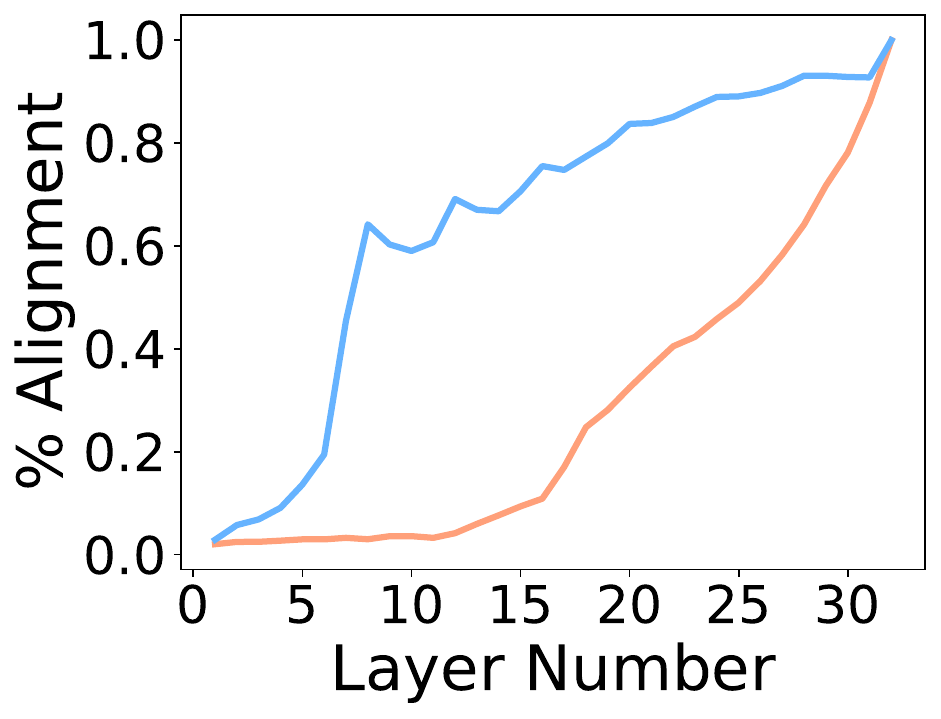}         
         \caption{Self-Instruct}        
    \end{subfigure}    
    \caption{Comparing percentage `alignment' of intermediate layer token predictions with the token predictions of the final layer for the model tuned using IT (orange) and the model tuned using IT with LITE (blue). This result is aggregated over all the output token predictions for all the inputs of the corresponding dataset.}
    \label{fig:token_alignment}    
\end{figure}

In Figure \ref{fig:token_alignment}, we plot the percentage alignment of token predictions of all intermediate layers with the token predictions of the final layer.
The figure shows the percentage alignment of (i) the model tuned using standard IT (orange) and (ii) the model tuned using IT with \texttt{LITE} (blue).
We show this result aggregated over all the output token predictions for all the inputs of the corresponding dataset.

We draw the following inferences: 

(a) \textbf{The predictions of the intermediate layers of the model tuned with LITE align well with the the final layer}, i.e., given a prefix for this model, the intermediate layers' token predictions match quite well with the final layer's token prediction.

(b) In contrast \textbf{for the model tuned using IT, the token predictions of the intermediate layers do not align that well with the token prediction of its final layer}.

(c) As the layer number increases, the percentage alignment also increases, i.e., \textbf{given a prefix, the predicted token of the later layers show higher alignment  (with predicted token of the final layer) than the initial layers}.

(d) There are \textbf{some peaks in the blue curve} (IT with \texttt{LITE}) which correspond to the selected layers from which the intermediate loss is aggregated during tuning, i.e., these layers show higher alignment. Conversely, layers from which the loss is not aggregated during tuning tend to show slightly lesser alignment.

In summary, this study demonstrates that \textbf{IT with LITE greatly aligns the token predictions of intermediate layers with that of the final layer}.

In the next subsection, we show that the token prediction probabilities of the intermediate layers provide a strong signal of this alignment, i.e., when the probability is high the token prediction of an intermediate layer is more likely to align with the token prediction of the final layer.
These two findings motivate dynamic confidence-based early exiting.

\subsection{In IT with \texttt{LITE}, Intermediate Layers' Token Prediction Probabilities Indicate Likelihood of Alignment with the Final Layer's Token Prediction}
\label{sec_confidence_alignment}

\begin{figure}[!]
\centering
    \begin{subfigure}{0.9\linewidth}
        \includegraphics[width=\linewidth]{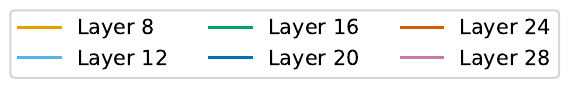}
    \end{subfigure}


    \begin{subfigure}{.48\linewidth}
        \includegraphics[width=\linewidth]{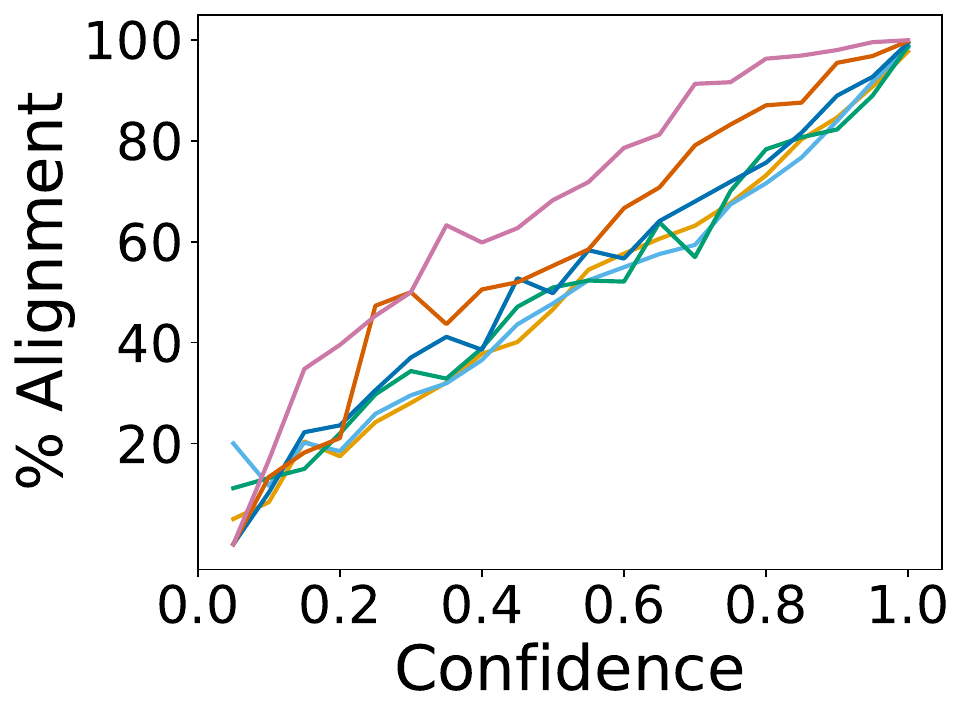}        
        \caption{Vicuna}        
    \end{subfigure}
    \begin{subfigure}{.48\linewidth}
         \includegraphics[width=\linewidth]{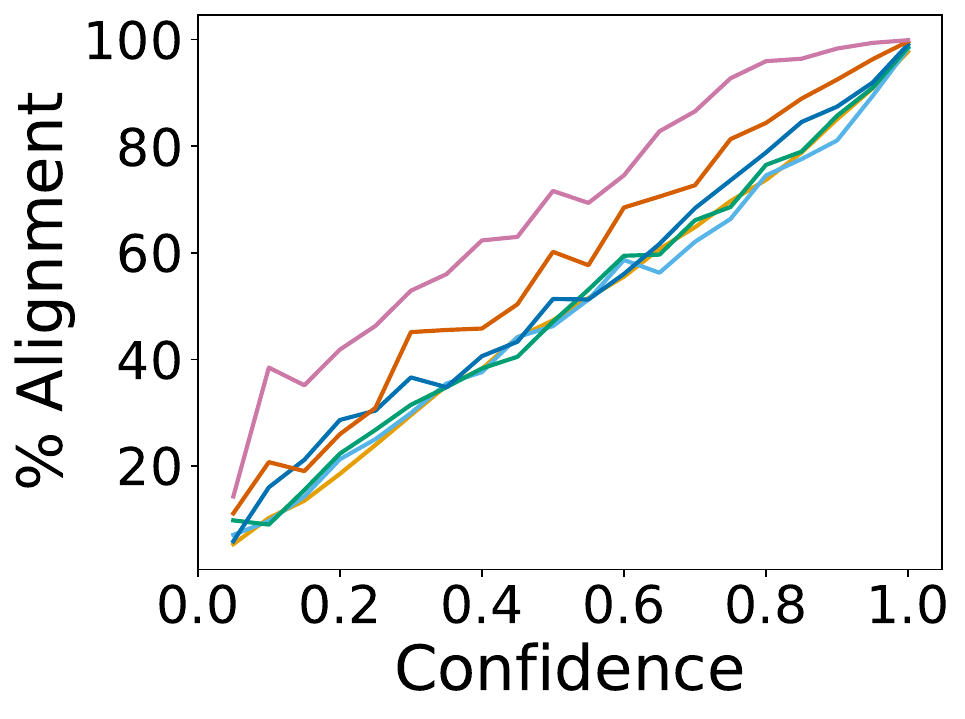}         
         \caption{Koala}
    \end{subfigure}   
    
    \begin{subfigure}{.48\linewidth}        
         \includegraphics[width=\linewidth]{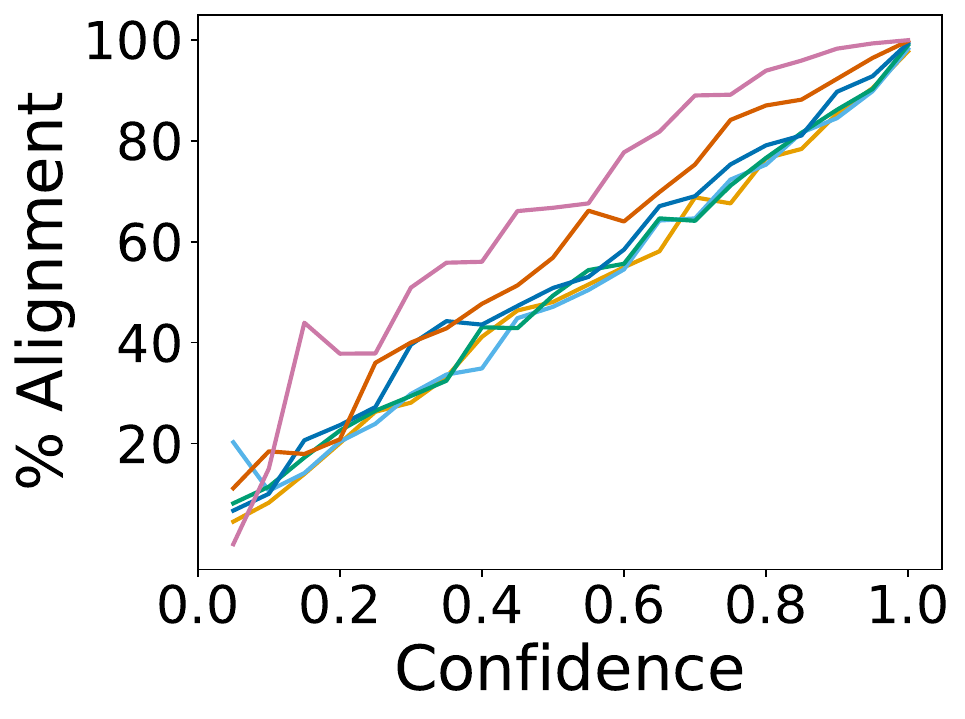}         
         \caption{WizardLM}
    \end{subfigure}    
    \begin{subfigure}{.48\linewidth}
         \includegraphics[width=\linewidth]{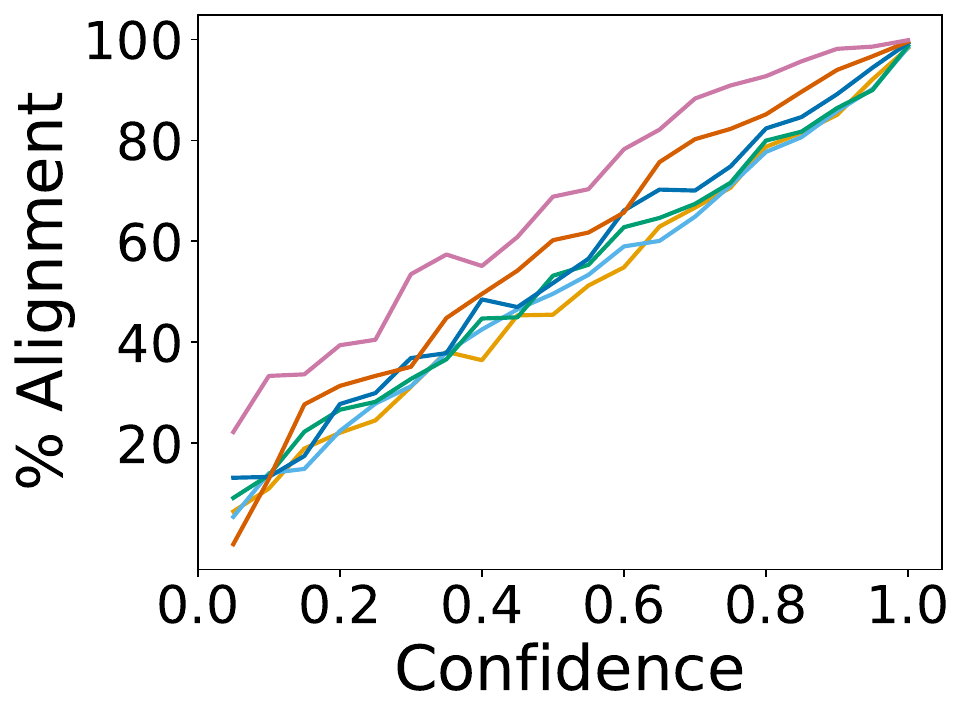}         
         \caption{Self-Instruct}        
    \end{subfigure}    
    \caption{Demonstrating relationship between token prediction confidence of the intermediate layers and the percentage alignment with the token prediction of the final layer for model tuned with \texttt{LITE}.
    \textbf{It shows that in IT with \texttt{LITE}, intermediate layers' token prediction probabilities (confidences) provide a strong signal of alignment with the final layer's token prediction.}
    }
    \label{fig:confidence_alignment}    
\end{figure}

Here, we plot the relationship between the token prediction confidence (softmax over the logits of the LM head) of the intermediate layers and the percentage alignment with the token prediction of the final layer.
Figure \ref{fig:confidence_alignment} shows this plot for the model tuned with \texttt{LITE}.
The figure shows that in IT with \texttt{LITE}, the intermediate layers' token prediction probabilities provide a strong signal of alignment, i.e., a high token prediction confidence implies a higher likelihood of its alignment with the token prediction of the final layer.
It also shows that with the increase in the layer number, the percentage alignment typically increases at the same confidence values. 

In contrast, in standard instruction tuning (IT), the confidence is not well correlated with the percentage alignment as we show in Appendix \ref{sec_confidence_alignment_IT}.
    
Building on the two findings (in \ref{sec_LITE_aligns_tokens} and \ref{sec_confidence_alignment}), \textbf{we perform `\textit{dynamic confidence-based early exiting}' at \underline{token level} from the intermediate layers and show that it improves the efficiency of inference while maintaining the generation quality}.

\subsection{Dynamic Confidence-Based Early Exiting Improves the Inference Efficiency While Maintaining the Generation Quality}
\label{sec_performance_early_exiting}

Motivated by the findings of the previous two subsections, we perform dynamic confidence-based early exiting at token-level, i.e., we exit when the token prediction confidence of the intermediate layer is sufficiently high (thus it is likely to align with the final layer's prediction).

To this end, from the confidence vs percentage alignment curve, we identify a confidence threshold for each layer where the alignment is $> 95\%$. 
Specifically, we use the following confidence thresholds:
Layer 8: $0.95$, 
Layer 12: $0.95$, 
Layer 16: $0.9$, 
Layer 20: $0.9$,
Layer 24: $0.8$, and
Layer 28: $0.7$.

In the main paper, we present the results and analysis for the aforementioned configuration.
However, we note that a different threshold configuration can also be used for inference. 
For instance, a more aggressive configuration with lower thresholds (shown in Appendix \ref{appendix_configuration}) leads to even more cost improvements ($49.92\%$); though it slightly drops the quality of generation ($5.34\%$). 
The trade-off between quality and cost can be balanced depending the application requirements.
For example, applications with quality tolerance or resource limitations can keep low threshold to achieve high cost improvements.

\textbf{Dynamic confidence-based early exiting: } At a selected layer, we pass its representations through the LM head,  calculate the softmax logit value, and compare it with the corresponding confidence threshold. 
If it surpasses the threshold value then we exit from that layer and proceed to generate the next token, otherwise we repeat this process at the next selected layer.

\begin{table}[t]
    \centering
    {
    \begin{tabular}{@{}cc@{}}
        \toprule
         \textbf{Test Dataset} & \textbf{Cost Improvement (\%)}\\
         
        \midrule
        
        Vicuna & 33.39 \% \\
        Koala & 35.40 \% \\
        WizardLM & 36.12 \%\\
        Self Instruct & 46.54 \% \\
        
    \bottomrule
    \end{tabular}    
    }
    \caption{
    Percentage improvements in the inference cost (measured in FLOPs) with dynamic early exiting. On average, it results in an improvement of 37.86\%.
    }

    \label{tab:percentage_improvements}
\end{table}

Figure \ref{fig:teaser} (in Section \ref{sec_introduction}) compares the \textbf{quality of responses} and the \textbf{inference cost} (measured in FLOPs) of the standard generation method (final layer) with our dynamic early exiting method.
It shows that \textbf{the dynamic early exiting method achieves consistent and considerable cost improvements (\bm{$37.86\%$} for 7B and \bm{$46.35\%$} for 13B model on average) while maintaining the generation quality}.
Table \ref{tab:percentage_improvements} shows the percentage improvements in inference cost for each test set individually.

We note that we use FLOPs as the metric of showcasing inference efficiency improvements because it is hardware independent, unlike latency.

In the remainder of this subsection, we present a thorough analysis of the results.
Specifically, we first present results at category level for the Vicuna and WizardLM datasets (\ref{sec_category_level_analysis}).
Then, to analyze the difference between the outputs of the two methods, we compare the semantic similarity between their responses (\ref{sec_semantic_similarity}).
Then, we dissect the effectiveness of the dynamic early exiting method in improving the computational efficiency by showing that both the methods result in comparable number of output tokens (\ref{sec_number_of_tokens}).
Finally, we show the percentage of exits from different layers in dynamic early exiting (\ref{sec_percentage_exits}).

\subsubsection{Quality and Inference Cost Analysis at Category Level}
\label{sec_category_level_analysis}

Vicuna and WizardLM test sets also provide the category corresponding to different test instances. 
To this end, we present category-level quality and inference cost results for these datasets.

\paragraph{Vicuna: }
Figure \ref{fig:vicuna_category_level_performance} (Appendix) compares the quality of responses and the inference cost of the standard generation method (final layer) with our dynamic early exiting method for different categories of Vicuna test set.
On average, it results in cost improvement of $33.39\%$.

\paragraph{WizardLM:}
Figure \ref{fig:wizardLM_category_level_performance}  (Appendix) compares the quality of responses and the inference cost of the standard generation method (final layer) with our dynamic early exiting method for different categories of WizardLM test set. 
On average, it results in cost improvement of $36.12\%$.

\subsubsection{
Dynamic Early Exiting Maintains the Semantics of the Responses}
\label{sec_semantic_similarity}

\begin{table}[t]
    \centering
    {
    \begin{tabular}{@{}cc@{}}
        \toprule
         \textbf{Test Dataset} & \textbf{Semantic Similarity}\\
         
        \midrule
        
        Vicuna & 0.9135 \\
        Koala & 0.8940 \\
        WizardLM & 0.9020 \\
        Self Instruct & 0.9001  \\
        
    \bottomrule
    \end{tabular}    
    }
    \caption{
    Semantic similarity between the final layer’s responses and the dynamic early exiting responses on the four test sets.
    Large similarity scores (closer to 1) imply that \textbf{dynamic early exiting maintains the semantics of the responses}.
    }
    \label{tab:semantic_similarity}
\end{table}

In addition to comparing the quality, we also compare the semantic similarity between the responses of the final layer and the dynamic early exiting.
Table \ref{tab:semantic_similarity} shows the semantic similarity (calculated using the `en\_core\_web\_sm' spacy model) for the four datasets.
It shows that there is a large semantic similarity between the responses as the values are closer to 1. 
This implies that dynamic early exiting maintains the semantics of the responses while providing efficiency benefits.
Appendix \ref{sec_response_comparison} shows examples of responses from both the last layer and our dynamic early exiting method.

\subsubsection{Dissecting the Improvements in Inference Cost}
\label{sec_number_of_tokens}

\begin{figure}[!]
    \centering
    \includegraphics[width=7cm]{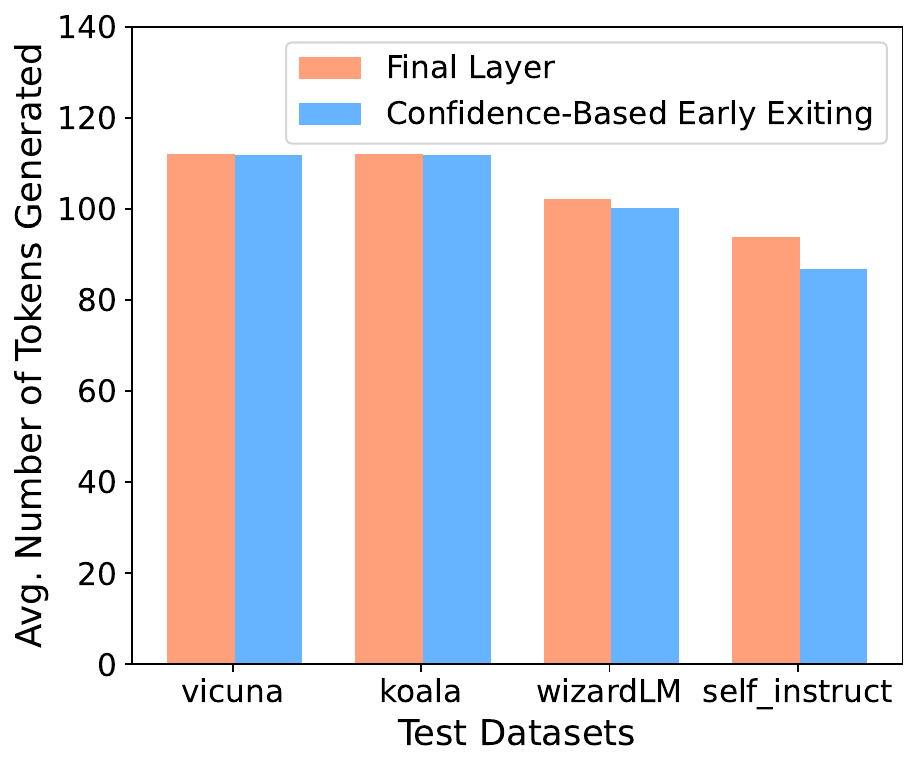}
    \caption{Comparing the average number of tokens generated in the final layer's responses and the dynamic early exiting responses for the four datasets.}    
    \label{fig:number_of_tokens_generated}
\end{figure}

In Figure \ref{fig:number_of_tokens_generated}, we compare the average number of tokens generated in the final layer's responses and the dynamic early exiting responses. 
It shows that both the methods generate a comparable number of tokens in their respective outputs. 
\textbf{This asserts that the cost improvement resulting in dynamic early exiting is because of the reduced computations due to early exiting and not due to generating a lesser number of tokens.}

\subsubsection{Percentage of Token Outputs from Different Exit Layers}
\label{sec_percentage_exits}

\begin{figure}[!]
    \centering
    \includegraphics[width=7cm]{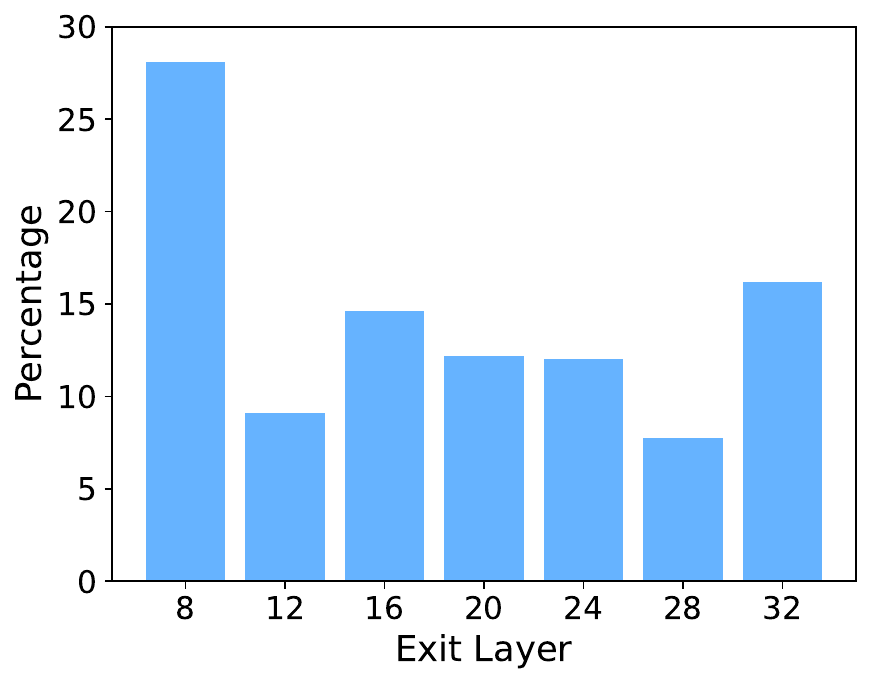}
    \caption{Percentage of token outputs from different exiting layers in the proposed method.}    
    \label{fig:exit_layers}
\end{figure}

Figure \ref{fig:exit_layers} shows the percentage of token outputs from different exit layers.
Note that this is aggregated across all the token positions.
This shows that the model exits a considerable percentage of times from the intermediate layers (while maintaining the generation quality) which further justifies the improvement in inference efficiency.

\textbf{We further conduct several interesting studies and analyses of the results and present them in Appendix \ref{sec_appendix_results}.}

\section{Conclusion and Discussion}
In this work, we proposed instruction tuning with additional explicit losses from the intermediate layers (\texttt{LITE}) and showed that it enables these layers to acquire `good' generation ability without affecting the final layer's generation ability.
We performed `\textit{dynamic confidence-based early exiting}' at {token level} from the intermediate layers and showed that it improves the efficiency of inference while maintaining the generation quality.
We conducted comprehensive experiments by instruction tuning LLaMA-2 models on the Alpaca dataset and evaluating on four different human-instruction test sets.
We showed that {dynamic early exiting achieves consistent and considerable inference cost improvements ({$37.86\%$} for 7B and {$46.35\%$} for 13B model) while maintaining the generation quality of the responses.
We further conducted a thorough analysis which resulted in several important findings.
Overall, our work contributes to improving the efficiency of LLM inference while maintaining the generation quality, a crucial step en route to enabling their widespread adoption.

Looking forward, our work additionally opens up several other avenues for new research, such as \textbf{speculative sampling from the intermediate layers} to improve the inference efficiency and \textbf{checking information consistency from the output of intermediate layers} to detect hallucinations.
Furthermore, this approach is complementary to some of the existing efficiency inference methods described in Section \ref{sec_related_work}, i.e., they can be used in a composite way to achieve even more efficiency gains.

\paragraph{Speculative Sampling: }
In speculative sampling \cite{leviathan2023fast, chen2023accelerating}, a smaller model is used as a drafting model. 
However, we showed that instruction tuning with LITE enables the intermediate layers to acquire `good' generation ability.
Thus, an intermediate layer of the same model can be used as the draft model while the last layer remains to be the target model.
This circumvents maintaining a separate drafting model for speculative sampling.
In this method, KV caching can also be used.

\paragraph{Hallucination Detection:}
Addressing the hallucination problem of LLMs is an important research direction and a number of methods have been developed \cite{varshney2023stitch,manakul2023selfcheckgpt,azaria2023internal,zhang2023interpretable,dhuliawala2023chain,gou2023critic}.
One of the popular methods requires generating multiple samples and then checking the information consistency between them. 
Here, we can use the intermediate layers to generate the output and then check the consistency between them.

\bibliography{anthology,custom}

\begin{thebibliography}{52}
\expandafter\ifx\csname natexlab\endcsname\relax\def\natexlab#1{#1}\fi

\bibitem[{Azaria and Mitchell(2023)}]{azaria2023internal}
Amos Azaria and Tom Mitchell. 2023.
\newblock The internal state of an llm knows when its lying.
\newblock \emph{arXiv preprint arXiv:2304.13734}.

\bibitem[{Bai et~al.(2022)Bai, Kadavath, Kundu, Askell, Kernion, Jones, Chen, Goldie, Mirhoseini, McKinnon et~al.}]{bai2022constitutional}
Yuntao Bai, Saurav Kadavath, Sandipan Kundu, Amanda Askell, Jackson Kernion, Andy Jones, Anna Chen, Anna Goldie, Azalia Mirhoseini, Cameron McKinnon, et~al. 2022.
\newblock Constitutional ai: Harmlessness from ai feedback.
\newblock \emph{arXiv preprint arXiv:2212.08073}.

\bibitem[{Brown et~al.(2020)Brown, Mann, Ryder, Subbiah, Kaplan, Dhariwal, Neelakantan, Shyam, Sastry, Askell, Agarwal, Herbert-Voss, Krueger, Henighan, Child, Ramesh, Ziegler, Wu, Winter, Hesse, Chen, Sigler, Litwin, Gray, Chess, Clark, Berner, McCandlish, Radford, Sutskever, and Amodei}]{NEURIPS2020_1457c0d6}
Tom Brown, Benjamin Mann, Nick Ryder, Melanie Subbiah, Jared~D Kaplan, Prafulla Dhariwal, Arvind Neelakantan, Pranav Shyam, Girish Sastry, Amanda Askell, Sandhini Agarwal, Ariel Herbert-Voss, Gretchen Krueger, Tom Henighan, Rewon Child, Aditya Ramesh, Daniel Ziegler, Jeffrey Wu, Clemens Winter, Chris Hesse, Mark Chen, Eric Sigler, Mateusz Litwin, Scott Gray, Benjamin Chess, Jack Clark, Christopher Berner, Sam McCandlish, Alec Radford, Ilya Sutskever, and Dario Amodei. 2020.
\newblock \href {https://proceedings.neurips.cc/paper_files/paper/2020/file/1457c0d6bfcb4967418bfb8ac142f64a-Paper.pdf} {Language models are few-shot learners}.
\newblock In \emph{Advances in Neural Information Processing Systems}, volume~33, pages 1877--1901. Curran Associates, Inc.

\bibitem[{Chen et~al.(2023)Chen, Borgeaud, Irving, Lespiau, Sifre, and Jumper}]{chen2023accelerating}
Charlie Chen, Sebastian Borgeaud, Geoffrey Irving, Jean-Baptiste Lespiau, Laurent Sifre, and John Jumper. 2023.
\newblock Accelerating large language model decoding with speculative sampling.
\newblock \emph{arXiv preprint arXiv:2302.01318}.

\bibitem[{Cheng et~al.(2023)Cheng, Kasai, and Yu}]{cheng2023batch}
Zhoujun Cheng, Jungo Kasai, and Tao Yu. 2023.
\newblock Batch prompting: Efficient inference with large language model apis.
\newblock \emph{arXiv preprint arXiv:2301.08721}.

\bibitem[{Chiang et~al.(2023)Chiang, Li, Lin, Sheng, Wu, Zhang, Zheng, Zhuang, Zhuang, Gonzalez, Stoica, and Xing}]{vicuna2023}
Wei-Lin Chiang, Zhuohan Li, Zi~Lin, Ying Sheng, Zhanghao Wu, Hao Zhang, Lianmin Zheng, Siyuan Zhuang, Yonghao Zhuang, Joseph~E. Gonzalez, Ion Stoica, and Eric~P. Xing. 2023.
\newblock \href {https://lmsys.org/blog/2023-03-30-vicuna/} {Vicuna: An open-source chatbot impressing gpt-4 with 90\%* chatgpt quality}.

\bibitem[{Chowdhery et~al.(2022)Chowdhery, Narang, Devlin, Bosma, Mishra, Roberts, Barham, Chung, Sutton, Gehrmann et~al.}]{chowdhery2022palm}
Aakanksha Chowdhery, Sharan Narang, Jacob Devlin, Maarten Bosma, Gaurav Mishra, Adam Roberts, Paul Barham, Hyung~Won Chung, Charles Sutton, Sebastian Gehrmann, et~al. 2022.
\newblock Palm: Scaling language modeling with pathways.
\newblock \emph{arXiv preprint arXiv:2204.02311}.

\bibitem[{Chung et~al.(2022)Chung, Hou, Longpre, Zoph, Tay, Fedus, Li, Wang, Dehghani, Brahma et~al.}]{chung2022scaling}
Hyung~Won Chung, Le~Hou, Shayne Longpre, Barret Zoph, Yi~Tay, William Fedus, Eric Li, Xuezhi Wang, Mostafa Dehghani, Siddhartha Brahma, et~al. 2022.
\newblock Scaling instruction-finetuned language models.
\newblock \emph{arXiv preprint arXiv:2210.11416}.

\bibitem[{Dettmers et~al.(2022)Dettmers, Lewis, Belkada, and Zettlemoyer}]{dettmers2022llm}
Tim Dettmers, Mike Lewis, Younes Belkada, and Luke Zettlemoyer. 2022.
\newblock Llm. int8 (): 8-bit matrix multiplication for transformers at scale.
\newblock \emph{arXiv preprint arXiv:2208.07339}.

\bibitem[{Dhuliawala et~al.(2023)Dhuliawala, Komeili, Xu, Raileanu, Li, Celikyilmaz, and Weston}]{dhuliawala2023chain}
Shehzaad Dhuliawala, Mojtaba Komeili, Jing Xu, Roberta Raileanu, Xian Li, Asli Celikyilmaz, and Jason Weston. 2023.
\newblock Chain-of-verification reduces hallucination in large language models.
\newblock \emph{arXiv preprint arXiv:2309.11495}.

\bibitem[{Din et~al.(2023)Din, Karidi, Choshen, and Geva}]{din2023jump}
Alexander~Yom Din, Taelin Karidi, Leshem Choshen, and Mor Geva. 2023.
\newblock Jump to conclusions: Short-cutting transformers with linear transformations.
\newblock \emph{arXiv preprint arXiv:2303.09435}.

\bibitem[{Elbayad et~al.(2020)Elbayad, Gu, Grave, and Auli}]{Elbayad2020Depth-Adaptive}
Maha Elbayad, Jiatao Gu, Edouard Grave, and Michael Auli. 2020.
\newblock \href {https://openreview.net/forum?id=SJg7KhVKPH} {Depth-adaptive transformer}.
\newblock In \emph{International Conference on Learning Representations}.

\bibitem[{Frantar et~al.(2022)Frantar, Ashkboos, Hoefler, and Alistarh}]{frantar2022gptq}
Elias Frantar, Saleh Ashkboos, Torsten Hoefler, and Dan Alistarh. 2022.
\newblock Gptq: Accurate post-training quantization for generative pre-trained transformers.
\newblock \emph{arXiv preprint arXiv:2210.17323}.

\bibitem[{Geng et~al.(2023)Geng, Gudibande, Liu, Wallace, Abbeel, Levine, and Song}]{koala_blogpost_2023}
Xinyang Geng, Arnav Gudibande, Hao Liu, Eric Wallace, Pieter Abbeel, Sergey Levine, and Dawn Song. 2023.
\newblock \href {https://bair.berkeley.edu/blog/2023/04/03/koala/} {Koala: A dialogue model for academic research}.
\newblock Blog post.

\bibitem[{Gera et~al.(2023)Gera, Friedman, Arviv, Gunasekara, Sznajder, Slonim, and Shnarch}]{gera2023benefits}
Ariel Gera, Roni Friedman, Ofir Arviv, Chulaka Gunasekara, Benjamin Sznajder, Noam Slonim, and Eyal Shnarch. 2023.
\newblock The benefits of bad advice: Autocontrastive decoding across model layers.
\newblock \emph{arXiv preprint arXiv:2305.01628}.

\bibitem[{Gou et~al.(2023)Gou, Shao, Gong, Shen, Yang, Duan, and Chen}]{gou2023critic}
Zhibin Gou, Zhihong Shao, Yeyun Gong, Yelong Shen, Yujiu Yang, Nan Duan, and Weizhu Chen. 2023.
\newblock Critic: Large language models can self-correct with tool-interactive critiquing.
\newblock \emph{arXiv preprint arXiv:2305.11738}.

\bibitem[{Goyal et~al.(2020)Goyal, Choudhury, Raje, Chakaravarthy, Sabharwal, and Verma}]{goyal2020power}
Saurabh Goyal, Anamitra~Roy Choudhury, Saurabh Raje, Venkatesan Chakaravarthy, Yogish Sabharwal, and Ashish Verma. 2020.
\newblock Power-bert: Accelerating bert inference via progressive word-vector elimination.
\newblock In \emph{International Conference on Machine Learning}, pages 3690--3699. PMLR.

\bibitem[{Guo et~al.(2021)Guo, Rush, and Kim}]{guo-etal-2021-parameter}
Demi Guo, Alexander Rush, and Yoon Kim. 2021.
\newblock \href {https://doi.org/10.18653/v1/2021.acl-long.378} {Parameter-efficient transfer learning with diff pruning}.
\newblock In \emph{Proceedings of the 59th Annual Meeting of the Association for Computational Linguistics and the 11th International Joint Conference on Natural Language Processing (Volume 1: Long Papers)}, pages 4884--4896, Online. Association for Computational Linguistics.

\bibitem[{Hou et~al.(2020)Hou, Huang, Shang, Jiang, Chen, and Liu}]{NEURIPS2020_6f5216f8}
Lu~Hou, Zhiqi Huang, Lifeng Shang, Xin Jiang, Xiao Chen, and Qun Liu. 2020.
\newblock \href {https://proceedings.neurips.cc/paper/2020/file/6f5216f8d89b086c18298e043bfe48ed-Paper.pdf} {Dynabert: Dynamic bert with adaptive width and depth}.
\newblock In \emph{Advances in Neural Information Processing Systems}, volume~33, pages 9782--9793. Curran Associates, Inc.

\bibitem[{Hsieh et~al.(2023)Hsieh, Li, Yeh, Nakhost, Fujii, Ratner, Krishna, Lee, and Pfister}]{hsieh2023distilling}
Cheng-Yu Hsieh, Chun-Liang Li, Chih-Kuan Yeh, Hootan Nakhost, Yasuhisa Fujii, Alexander Ratner, Ranjay Krishna, Chen-Yu Lee, and Tomas Pfister. 2023.
\newblock Distilling step-by-step! outperforming larger language models with less training data and smaller model sizes.
\newblock \emph{arXiv preprint arXiv:2305.02301}.

\bibitem[{Jiao et~al.(2020)Jiao, Yin, Shang, Jiang, Chen, Li, Wang, and Liu}]{jiao-etal-2020-tinybert}
Xiaoqi Jiao, Yichun Yin, Lifeng Shang, Xin Jiang, Xiao Chen, Linlin Li, Fang Wang, and Qun Liu. 2020.
\newblock \href {https://doi.org/10.18653/v1/2020.findings-emnlp.372} {{T}iny{BERT}: Distilling {BERT} for natural language understanding}.
\newblock In \emph{Findings of the Association for Computational Linguistics: EMNLP 2020}, pages 4163--4174, Online. Association for Computational Linguistics.

\bibitem[{Leviathan et~al.(2023)Leviathan, Kalman, and Matias}]{leviathan2023fast}
Yaniv Leviathan, Matan Kalman, and Yossi Matias. 2023.
\newblock Fast inference from transformers via speculative decoding.
\newblock In \emph{International Conference on Machine Learning}, pages 19274--19286. PMLR.

\bibitem[{Li et~al.(2021)Li, Lin, Chen, Ren, Li, Zhou, and Sun}]{li-etal-2021-cascadebert-accelerating}
Lei Li, Yankai Lin, Deli Chen, Shuhuai Ren, Peng Li, Jie Zhou, and Xu~Sun. 2021.
\newblock \href {https://doi.org/10.18653/v1/2021.findings-emnlp.43} {{C}ascade{BERT}: Accelerating inference of pre-trained language models via calibrated complete models cascade}.
\newblock In \emph{Findings of the Association for Computational Linguistics: EMNLP 2021}, pages 475--486, Punta Cana, Dominican Republic. Association for Computational Linguistics.

\bibitem[{Li et~al.(2022)Li, Wang, Tan, Nallapati, Bhatia, Arnold, Xiang, and Roth}]{li-etal-2022-dq}
Zheng Li, Zijian Wang, Ming Tan, Ramesh Nallapati, Parminder Bhatia, Andrew Arnold, Bing Xiang, and Dan Roth. 2022.
\newblock \href {https://doi.org/10.18653/v1/2022.acl-short.22} {{DQ}-{BART}: Efficient sequence-to-sequence model via joint distillation and quantization}.
\newblock In \emph{Proceedings of the 60th Annual Meeting of the Association for Computational Linguistics (Volume 2: Short Papers)}, pages 203--211, Dublin, Ireland. Association for Computational Linguistics.

\bibitem[{Manakul et~al.(2023)Manakul, Liusie, and Gales}]{manakul2023selfcheckgpt}
Potsawee Manakul, Adian Liusie, and Mark~JF Gales. 2023.
\newblock Selfcheckgpt: Zero-resource black-box hallucination detection for generative large language models.
\newblock \emph{arXiv preprint arXiv:2303.08896}.

\bibitem[{Mirzadeh et~al.(2020)Mirzadeh, Farajtabar, Li, Levine, Matsukawa, and Ghasemzadeh}]{mirzadeh2020improved}
Seyed~Iman Mirzadeh, Mehrdad Farajtabar, Ang Li, Nir Levine, Akihiro Matsukawa, and Hassan Ghasemzadeh. 2020.
\newblock Improved knowledge distillation via teacher assistant.
\newblock In \emph{Proceedings of the AAAI Conference on Artificial Intelligence}, volume~34, pages 5191--5198.

\bibitem[{Mishra et~al.(2022)Mishra, Khashabi, Baral, and Hajishirzi}]{mishra-etal-2022-cross}
Swaroop Mishra, Daniel Khashabi, Chitta Baral, and Hannaneh Hajishirzi. 2022.
\newblock \href {https://doi.org/10.18653/v1/2022.acl-long.244} {Cross-task generalization via natural language crowdsourcing instructions}.
\newblock In \emph{Proceedings of the 60th Annual Meeting of the Association for Computational Linguistics (Volume 1: Long Papers)}, pages 3470--3487, Dublin, Ireland. Association for Computational Linguistics.

\bibitem[{O'Brien and Lewis(2023)}]{o2023contrastive}
Sean O'Brien and Mike Lewis. 2023.
\newblock Contrastive decoding improves reasoning in large language models.
\newblock \emph{arXiv preprint arXiv:2309.09117}.

\bibitem[{OpenAI(2023)}]{OpenAI2023GPT4TR}
OpenAI. 2023.
\newblock \href {https://api.semanticscholar.org/CorpusID:257532815} {Gpt-4 technical report}.
\newblock \emph{ArXiv}, abs/2303.08774.

\bibitem[{Parmar et~al.(2022)Parmar, Mishra, Purohit, Luo, Mohammad, and Baral}]{parmar-etal-2022-boxbart}
Mihir Parmar, Swaroop Mishra, Mirali Purohit, Man Luo, Murad Mohammad, and Chitta Baral. 2022.
\newblock \href {https://doi.org/10.18653/v1/2022.findings-naacl.10} {In-{B}o{XBART}: Get instructions into biomedical multi-task learning}.
\newblock In \emph{Findings of the Association for Computational Linguistics: NAACL 2022}, pages 112--128, Seattle, United States. Association for Computational Linguistics.

\bibitem[{Rae et~al.(2021)Rae, Borgeaud, Cai, Millican, Hoffmann, Song, Aslanides, Henderson, Ring, Young et~al.}]{rae2021scaling}
Jack~W Rae, Sebastian Borgeaud, Trevor Cai, Katie Millican, Jordan Hoffmann, Francis Song, John Aslanides, Sarah Henderson, Roman Ring, Susannah Young, et~al. 2021.
\newblock Scaling language models: Methods, analysis \& insights from training gopher.
\newblock \emph{arXiv preprint arXiv:2112.11446}.

\bibitem[{Sanh et~al.(2022)Sanh, Webson, Raffel, Bach, Sutawika, Alyafeai, Chaffin, Stiegler, Raja, Dey, Bari, Xu, Thakker, Sharma, Szczechla, Kim, Chhablani, Nayak, Datta, Chang, Jiang, Wang, Manica, Shen, Yong, Pandey, Bawden, Wang, Neeraj, Rozen, Sharma, Santilli, Fevry, Fries, Teehan, Scao, Biderman, Gao, Wolf, and Rush}]{sanh2022multitask}
Victor Sanh, Albert Webson, Colin Raffel, Stephen Bach, Lintang Sutawika, Zaid Alyafeai, Antoine Chaffin, Arnaud Stiegler, Arun Raja, Manan Dey, M~Saiful Bari, Canwen Xu, Urmish Thakker, Shanya~Sharma Sharma, Eliza Szczechla, Taewoon Kim, Gunjan Chhablani, Nihal Nayak, Debajyoti Datta, Jonathan Chang, Mike Tian-Jian Jiang, Han Wang, Matteo Manica, Sheng Shen, Zheng~Xin Yong, Harshit Pandey, Rachel Bawden, Thomas Wang, Trishala Neeraj, Jos Rozen, Abheesht Sharma, Andrea Santilli, Thibault Fevry, Jason~Alan Fries, Ryan Teehan, Teven~Le Scao, Stella Biderman, Leo Gao, Thomas Wolf, and Alexander~M Rush. 2022.
\newblock \href {https://openreview.net/forum?id=9Vrb9D0WI4} {Multitask prompted training enables zero-shot task generalization}.
\newblock In \emph{International Conference on Learning Representations}.

\bibitem[{Schuster et~al.(2022)Schuster, Fisch, Gupta, Dehghani, Bahri, Tran, Tay, and Metzler}]{schuster2022confident}
Tal Schuster, Adam Fisch, Jai Gupta, Mostafa Dehghani, Dara Bahri, Vinh Tran, Yi~Tay, and Donald Metzler. 2022.
\newblock Confident adaptive language modeling.
\newblock \emph{Advances in Neural Information Processing Systems}, 35:17456--17472.

\bibitem[{Smith et~al.(2022)Smith, Patwary, Norick, LeGresley, Rajbhandari, Casper, Liu, Prabhumoye, Zerveas, Korthikanti et~al.}]{smith2022using}
Shaden Smith, Mostofa Patwary, Brandon Norick, Patrick LeGresley, Samyam Rajbhandari, Jared Casper, Zhun Liu, Shrimai Prabhumoye, George Zerveas, Vijay Korthikanti, et~al. 2022.
\newblock Using deepspeed and megatron to train megatron-turing nlg 530b, a large-scale generative language model.
\newblock \emph{arXiv preprint arXiv:2201.11990}.

\bibitem[{Taori et~al.(2023)Taori, Gulrajani, Zhang, Dubois, Li, Guestrin, Liang, and Hashimoto}]{alpaca}
Rohan Taori, Ishaan Gulrajani, Tianyi Zhang, Yann Dubois, Xuechen Li, Carlos Guestrin, Percy Liang, and Tatsunori~B. Hashimoto. 2023.
\newblock Stanford alpaca: An instruction-following llama model.
\newblock \url{https://github.com/tatsu-lab/stanford_alpaca}.

\bibitem[{Touvron et~al.(2023)Touvron, Martin, Stone, Albert, Almahairi, Babaei, Bashlykov, Batra, Bhargava, Bhosale et~al.}]{touvron2023llama}
Hugo Touvron, Louis Martin, Kevin Stone, Peter Albert, Amjad Almahairi, Yasmine Babaei, Nikolay Bashlykov, Soumya Batra, Prajjwal Bhargava, Shruti Bhosale, et~al. 2023.
\newblock Llama 2: Open foundation and fine-tuned chat models.
\newblock \emph{arXiv preprint arXiv:2307.09288}.

\bibitem[{Varshney and Baral(2022)}]{varshney-baral-2022-model}
Neeraj Varshney and Chitta Baral. 2022.
\newblock \href {https://doi.org/10.18653/v1/2022.emnlp-main.756} {Model cascading: Towards jointly improving efficiency and accuracy of {NLP} systems}.
\newblock In \emph{Proceedings of the 2022 Conference on Empirical Methods in Natural Language Processing}, pages 11007--11021, Abu Dhabi, United Arab Emirates. Association for Computational Linguistics.

\bibitem[{Varshney and Baral(2023)}]{varshney-baral-2023-post}
Neeraj Varshney and Chitta Baral. 2023.
\newblock \href {https://doi.org/10.18653/v1/2023.acl-long.55} {Post-abstention: Towards reliably re-attempting the abstained instances in {QA}}.
\newblock In \emph{Proceedings of the 61st Annual Meeting of the Association for Computational Linguistics (Volume 1: Long Papers)}, pages 967--982, Toronto, Canada. Association for Computational Linguistics.

\bibitem[{Varshney et~al.(2023)Varshney, Yao, Zhang, Chen, and Yu}]{varshney2023stitch}
Neeraj Varshney, Wenlin Yao, Hongming Zhang, Jianshu Chen, and Dong Yu. 2023.
\newblock A stitch in time saves nine: Detecting and mitigating hallucinations of llms by validating low-confidence generation.
\newblock \emph{arXiv preprint arXiv:2307.03987}.

\bibitem[{Wang et~al.(2023)Wang, Li, Chen, Zhu, Lin, Cao, Liu, Liu, and Sui}]{wang2023large}
Peiyi Wang, Lei Li, Liang Chen, Dawei Zhu, Binghuai Lin, Yunbo Cao, Qi~Liu, Tianyu Liu, and Zhifang Sui. 2023.
\newblock Large language models are not fair evaluators.
\newblock \emph{arXiv preprint arXiv:2305.17926}.

\bibitem[{Wang et~al.(2022{\natexlab{a}})Wang, Kordi, Mishra, Liu, Smith, Khashabi, and Hajishirzi}]{wang2022self}
Yizhong Wang, Yeganeh Kordi, Swaroop Mishra, Alisa Liu, Noah~A Smith, Daniel Khashabi, and Hannaneh Hajishirzi. 2022{\natexlab{a}}.
\newblock Self-instruct: Aligning language model with self generated instructions.
\newblock \emph{arXiv preprint arXiv:2212.10560}.

\bibitem[{Wang et~al.(2022{\natexlab{b}})Wang, Mishra, Alipoormolabashi, Kordi, Mirzaei, Naik, Ashok, Dhanasekaran, Arunkumar, Stap, Pathak, Karamanolakis, Lai, Purohit, Mondal, Anderson, Kuznia, Doshi, Pal, Patel, Moradshahi, Parmar, Purohit, Varshney, Kaza, Verma, Puri, Karia, Doshi, Sampat, Mishra, Reddy~A, Patro, Dixit, and Shen}]{wang-etal-2022-super}
Yizhong Wang, Swaroop Mishra, Pegah Alipoormolabashi, Yeganeh Kordi, Amirreza Mirzaei, Atharva Naik, Arjun Ashok, Arut~Selvan Dhanasekaran, Anjana Arunkumar, David Stap, Eshaan Pathak, Giannis Karamanolakis, Haizhi Lai, Ishan Purohit, Ishani Mondal, Jacob Anderson, Kirby Kuznia, Krima Doshi, Kuntal~Kumar Pal, Maitreya Patel, Mehrad Moradshahi, Mihir Parmar, Mirali Purohit, Neeraj Varshney, Phani~Rohitha Kaza, Pulkit Verma, Ravsehaj~Singh Puri, Rushang Karia, Savan Doshi, Shailaja~Keyur Sampat, Siddhartha Mishra, Sujan Reddy~A, Sumanta Patro, Tanay Dixit, and Xudong Shen. 2022{\natexlab{b}}.
\newblock \href {https://aclanthology.org/2022.emnlp-main.340} {Super-{N}atural{I}nstructions: Generalization via declarative instructions on 1600+ {NLP} tasks}.
\newblock In \emph{Proceedings of the 2022 Conference on Empirical Methods in Natural Language Processing}, pages 5085--5109, Abu Dhabi, United Arab Emirates. Association for Computational Linguistics.

\bibitem[{Wang et~al.(2020)Wang, Wohlwend, and Lei}]{wang-etal-2020-structured}
Ziheng Wang, Jeremy Wohlwend, and Tao Lei. 2020.
\newblock \href {https://doi.org/10.18653/v1/2020.emnlp-main.496} {Structured pruning of large language models}.
\newblock In \emph{Proceedings of the 2020 Conference on Empirical Methods in Natural Language Processing (EMNLP)}, pages 6151--6162, Online. Association for Computational Linguistics.

\bibitem[{Wei et~al.(2022)Wei, Bosma, Zhao, Guu, Yu, Lester, Du, Dai, and Le}]{wei2022finetuned}
Jason Wei, Maarten Bosma, Vincent Zhao, Kelvin Guu, Adams~Wei Yu, Brian Lester, Nan Du, Andrew~M. Dai, and Quoc~V Le. 2022.
\newblock \href {https://openreview.net/forum?id=gEZrGCozdqR} {Finetuned language models are zero-shot learners}.
\newblock In \emph{International Conference on Learning Representations}.

\bibitem[{Wei et~al.(2021)Wei, Bosma, Zhao, Guu, Yu, Lester, Du, Dai, and Le}]{wei2021finetuned}
Jason Wei, Maarten Bosma, Vincent~Y Zhao, Kelvin Guu, Adams~Wei Yu, Brian Lester, Nan Du, Andrew~M Dai, and Quoc~V Le. 2021.
\newblock Finetuned language models are zero-shot learners.
\newblock \emph{arXiv preprint arXiv:2109.01652}.

\bibitem[{Xiao et~al.(2023)Xiao, Lin, Seznec, Wu, Demouth, and Han}]{pmlr-v202-xiao23c}
Guangxuan Xiao, Ji~Lin, Mickael Seznec, Hao Wu, Julien Demouth, and Song Han. 2023.
\newblock \href {https://proceedings.mlr.press/v202/xiao23c.html} {{S}mooth{Q}uant: Accurate and efficient post-training quantization for large language models}.
\newblock In \emph{Proceedings of the 40th International Conference on Machine Learning}, volume 202 of \emph{Proceedings of Machine Learning Research}, pages 38087--38099. PMLR.

\bibitem[{Xin et~al.(2020)Xin, Tang, Lee, Yu, and Lin}]{xin-etal-2020-deebert}
Ji~Xin, Raphael Tang, Jaejun Lee, Yaoliang Yu, and Jimmy Lin. 2020.
\newblock \href {https://doi.org/10.18653/v1/2020.acl-main.204} {{D}ee{BERT}: Dynamic early exiting for accelerating {BERT} inference}.
\newblock In \emph{Proceedings of the 58th Annual Meeting of the Association for Computational Linguistics}, pages 2246--2251, Online. Association for Computational Linguistics.

\bibitem[{Xu et~al.(2023)Xu, Sun, Zheng, Geng, Zhao, Feng, Tao, and Jiang}]{xu2023wizardlm}
Can Xu, Qingfeng Sun, Kai Zheng, Xiubo Geng, Pu~Zhao, Jiazhan Feng, Chongyang Tao, and Daxin Jiang. 2023.
\newblock Wizardlm: Empowering large language models to follow complex instructions.
\newblock \emph{arXiv preprint arXiv:2304.12244}.

\bibitem[{Yao et~al.(2022)Yao, Yazdani~Aminabadi, Zhang, Wu, Li, and He}]{yao2022zeroquant}
Zhewei Yao, Reza Yazdani~Aminabadi, Minjia Zhang, Xiaoxia Wu, Conglong Li, and Yuxiong He. 2022.
\newblock Zeroquant: Efficient and affordable post-training quantization for large-scale transformers.
\newblock \emph{Advances in Neural Information Processing Systems}, 35:27168--27183.

\bibitem[{Yue et~al.(2023)Yue, Zhao, Zhang, Du, and Yao}]{yue2023large}
Murong Yue, Jie Zhao, Min Zhang, Liang Du, and Ziyu Yao. 2023.
\newblock Large language model cascades with mixture of thoughts representations for cost-efficient reasoning.
\newblock \emph{arXiv preprint arXiv:2310.03094}.

\bibitem[{Zhang et~al.(2023)Zhang, Luo, Chuang, Fang, Gaitskell, Hartvigsen, Wu, Fox, Meng, and Glass}]{zhang2023interpretable}
Tianhua Zhang, Hongyin Luo, Yung-Sung Chuang, Wei Fang, Luc Gaitskell, Thomas Hartvigsen, Xixin Wu, Danny Fox, Helen Meng, and James Glass. 2023.
\newblock Interpretable unified language checking.
\newblock \emph{arXiv preprint arXiv:2304.03728}.

\bibitem[{Zheng et~al.(2023)Zheng, Chiang, Sheng, Zhuang, Wu, Zhuang, Lin, Li, Li, Xing et~al.}]{zheng2023judging}
Lianmin Zheng, Wei-Lin Chiang, Ying Sheng, Siyuan Zhuang, Zhanghao Wu, Yonghao Zhuang, Zi~Lin, Zhuohan Li, Dacheng Li, Eric Xing, et~al. 2023.
\newblock Judging llm-as-a-judge with mt-bench and chatbot arena.
\newblock \emph{arXiv preprint arXiv:2306.05685}.

\end{thebibliography}
\bibliographystyle{acl_natbib}

\appendix
\newpage
\section*{Appendix}

\section{Evaluation Methodology}
\label{evaluation_prompt}
We use the following prompt with Claude model for comparing the responses of two models:
\begin{tcolorbox}
    \texttt{\textbf{Human}: You are a helpful and precise assistant for checking the quality of the answer.} \\ \\
    
    \texttt{\textbf{[Question]}
    \newline
    \{{question}\}
    \newline \newline
    \textbf{[The Start of Assistant 1's Answer]}
    \newline
    \{answer\_1\}
    \newline \newline
    \textbf{[The End of Assistant 1's Answer]}
    \newline \newline
    \textbf{[The Start of Assistant 2's Answer]}
    \newline
    \{answer\_2\}
    \newline \newline
    \textbf{[The End of Assistant 2's Answer]}
    \newline \newline
    We would like to request your feedback on the performance of two AI assistants in response to the user question displayed above.
    \newline 
    Please rate the helpfulness, relevance, accuracy, level of details of their responses. Each assistant receives an overall score on a scale of 1 to 10, where a higher score indicates better overall performance.
    \newline 
    Please output a single line containing only two values indicating the scores for Assistant 1 and 2, respectively. The two scores are separated by a space.
    \newline \newline
    \textbf{Assistant:}
    } 
    
\end{tcolorbox}

We use amazon bedrock to call the Claude API with 
temperature set to 0 and modelId to `anthropic.claude-v1'.

\section{Additional Results}
\label{sec_appendix_results}

\subsection{Inability of the Intermediate Layers of the Model Tuned with Standard Instruction Tuning to Generate `High-Quality' Text}
\label{sec_appendix_inability_of_intermediate}

In this subsection, we show the responses of intermediate layers (fixed exiting) of the model tuned with standard instruction tuning (IT) for a few examples.

\begin{tcolorbox}
    \texttt{\textbf{Question}: What are the differences between plant-based and animal-based protein sources?}\\ \\    
    \texttt{
    {\textbf{Layer 32 (Final)}}: 
    The main difference between plant-based and animal-based protein sources is that plant-based proteins are found in plant-based foods, such as legumes, nuts, and seeds, while animal-based proteins are found in animal-based foods, such as meat, eggs, and fish. Plant-based proteins typically contain fewer calories, fat, and cholesterol than animal-based proteins. Additionally, ... \\ \\
    \textbf{Layer 28:} 
    plant- based proteins sources such as legumes, legumes, and nuts contain essential amino achren needed for growth and repair of cellsular structures. animal based proteins sources such as meat, eggs, and fish contain additional nutrients such as vitamin B Complex vitamin D vitamin Iron and calcium needed for optimal health.  \\ \\
    \textbf{Layer 24:} 
    plant based prote source protein source diversity, quality proteinsrichness quantityquantityquantityquantity quantity quantity quantity quantity quantity q ...   \\ \\
    \textbf{Layer 20:} 
    difference between Plant Based vs animal based protein sourceage oldenhamptonensis proportionalityunicíclesqueque ... 
    } 
    
\end{tcolorbox}

\begin{tcolorbox}
    \texttt{\textbf{Question}: Imagine you are a time traveler from the year 3000. What technological advancements would you tell people about?}\\ \\    
    \texttt{
    {\textbf{Layer 32 (Final)}}: 
    From the year 3000, I would tell people about the advancements in technology that have made life easier and more efficient. I would talk about the autonomous cars, the implanted chips that allow people to access the internet and doctors to monitor patients remotely, the 3D-printed homes and the list goes on. I would also tell people about the advancements in renewable energy and how we are able to generate electricity from the sun and the wind.  \\ \\
    \textbf{Layer 28:} 
    From the year 3000 perspective, some technological advances might include flying cars, robotic vac vacuifiers, implanted implants used for implantable devices such as pacemakers, implanted cameras implanted microscopes implanted anything else you can think of implanted implanted implanted ... \\ \\
    \textbf{Layer 24:} 
    techn adv advances techn advances techn advances techn ...   \\ \\
    \textbf{Layer 20:} 
    futureproofing@gmailgmail domainainsights intoxicatednessipag eantryonswardswardsonswardsonswardso ... 
    } 
    
\end{tcolorbox}

The above examples and the results in Section \ref{sec_inability_of_regular_LLM} show that in standard instruction tuning, the intermediate layers do not acquire good generation ability. 
In the next subsection, we illustrate examples that support our argument that IT with LITE addresses the above problem.

\subsection{Intermediate Layers Acquire `Good' Generation Ability in IT with LITE }

\begin{tcolorbox}
    \texttt{\textbf{Question}: What are the differences between plant-based and animal-based protein sources?}\\ \\    
    \texttt{
    {\textbf{Layer 32 (Final)}}: 
    Plant-based protein sources are mainly found in foods such as legumes, nuts, and seeds, while animal-based protein sources are found in foods such as meat, fish, eggs, and dairy products. Plant-based proteins are generally lower in calories and fat, and are rich in vitamins, minerals, and fiber. Animal-based proteins are typically higher in calories and fat, and ... \\ \\
    \textbf{Layer 28:} 
    Plant-based protein sources are mainly found in foods such as legumes, nuts, and seeds, while animal-based protein sources are found in foods such as meat, fish, eggs, and dairy products. Plant-based proteins are generally lower in calories and fat, and are rich in vitamins, minerals, and fiber. Animal-based proteins are typically higher in calories  ...  \\ \\
    \textbf{Layer 24:} 
    Plant-based protein sources are mainly found in foods such as legumes, nuts, and seeds, and are typically lower in calories and fat than animal-based protein sources. Plant-based proteins are also rich in vitamins, minerals, and antioxidants, and are a good source of  ...   \\ \\
    \textbf{Layer 20:} 
    Plant-based protein sources are found in many different foods, including legumes, nuts, seeds, and seeds. They are usually lower in calories and fat than animal-based protein sources. Plant-based proteins are also rich in fiber and contain essential vitamins ... 
    } 
    
\end{tcolorbox}

\begin{tcolorbox}
    \texttt{\textbf{Question}: Imagine you are a time traveler from the year 3000. What technological advancements would you tell people about?}\\ \\    
    \texttt{
    {\textbf{Layer 32 (Final)}}: 
    From the year 3000, I would tell people about the incredible advancements in technology that have been made over the centuries. We take many things for granted now, such as the internet and autonomous vehicles, but these technologies were first developed in the  ... \\ \\
    \textbf{Layer 28:} 
    From the year 3000, I would tell people about the advancements in artificial intelligence, robotics, and genetic engineering. In particular, I would tell people about the advancements made in AI, such as the development of autonomous cars, virtual assistants, and smart homes. I would  ... \\ \\
    \textbf{Layer 24:} 
    From the year 3000, I would tell people about the incredible advancements in technology that have been made over the centuries. We are now able to control our environment by using thought-controlled prostheses, use genetic  ...   \\ \\
    \textbf{Layer 20:} 
    In the 3000s, we have developed advanced artificial intelligence, self-driving cars, and teleportation of particles. We have also made significant progress in quantum computing, which has allowed us to ...
    } 
    
\end{tcolorbox}

The above examples illustrate that IT with LITE enables the intermediate layers to acquire good generation ability.

\subsection{Poor Quality of Responses With Dynamic Early Exiting from Model Tuned with Standard Instruction Tuning (IT)}
\label{sec_response_comparison_IT}

We show examples of the responses generated via dynamic confidence-based early exiting from the model tuned with standard instruction tuning (IT \textbf{without} \texttt{LITE}).
These examples show that the early exiting responses are of poor quality.
They suffer from a number of issues including degeneration such as unwanted token repetitions, lack of coherence and structure.  

This showcases the importance of \texttt{LITE} during instruction tuning as it allows the intermediate layers to acquire good generation ability and thus enabling opportunities of early exiting without compromising the quality.

\begin{tcolorbox}
    \texttt{\textbf{Question}: What are the most effective ways to deal with stress?}\\ \\    
    \texttt{
    {\textbf{Layer 32 (Final)}}: 
    The most effective ways to deal with stress include identifying sources of stress, developing coping strategies, maintaining a balanced diet, exercising regularly, getting enough sleep, and taking time to relax and do something enjoyable. Taking time to practice deep breathing and meditation, talking to a friend or therapist, and using humor are also helpful. \\ \\
    \textbf{Dynamic Early Exiting with IT model:} 
     cop coping strategies include identifying sources of stress, developing cop coping skills such as cop coping and relaxation techniques such as meditation and med meditation med meditation med meditation med meditation med meditation med meditation med meditation med meditation med meditation ...\\
    } 
    
\end{tcolorbox}

\begin{tcolorbox}
    \texttt{\textbf{Question}: Can you explain the basics of quantum computing?}\\ \\    
    \texttt{
    {\textbf{Layer 32 (Final)}}: 
    Quantum computing is a form of computing that uses quantum-mechanical phenomena, such as superposition and entanglement, to perform operations on data. It enables computation at a much faster rate than traditional computers and can solve problems that are intractable on classical computers. Quantum computing uses qubits\u2014quantum bits\u2014which can be in a state of 0 or 1, instead of the traditional bit of 0 or 1. Qubits can be used to represent data and perform operations on it, such as addition, subtraction, and multiplication. \\ \\
    \textbf{Dynamic Early Exiting with IT model:} 
     quantum computing basics quantum computing basics explain the basics of quantum computing including what it is, how it works, why it matter, how it can be used for computing, and what its future may hold. quantum computing basics quantum computing basics explain the basics of quantum computing including what it is, how it works, why it matter, how it can be used for computing ... \\ 
    } 
    
\end{tcolorbox}

\subsection{Examples of Responses with Dynamic Early Exiting for Model instruction tuned with LITE}
\label{sec_response_comparison}

Here, we show show responses of the final layer and the dynamic early exiting method with the model instruction tuned with \texttt{LITE}.

\begin{tcolorbox}
    \texttt{\textbf{Question}: What are the main differences between Python and JavaScript programming languages?}\\ \\    
    \texttt{
    {\textbf{Layer 32 (Final)}}: 
    The main differences between Python and JavaScript programming languages are their syntax, data types, data structures, and logic. Python is a high-level, interpreted language that is focused on readability and simplicity, while JavaScript is a low-level, interpreted language that is focused on adding interactive elements to web pages. Python has strong support for data typing, while JavaScript has weaker support for data types. Python is more suitable for creating scripts and applications, while JavaScript is more suitable for creating interactive web pages. \\ \\
    \textbf{Dynamic Early Exiting:} 
    The main differences between Python and JavaScript programming languages are their syntax, data types, data structures, and logic. Python is a high-level, interpreted language that is focused on readability and simplicity, while JavaScript is a low-level, interpreted language that is focused on adding interactive features to web pages. Python has dynamic data types, while JavaScript has static data types. Python is more suitable for creating scripts and applications, while JavaScript is more suitable for creating interactive web pages.\\
    } 
    
\end{tcolorbox}


\begin{tcolorbox}
    \texttt{\textbf{Question}: What if Isaac Newton had focused on biology instead of physics?}\\ \\    
    \texttt{
    {\textbf{Layer 32 (Final)}}: 
    Isaac Newton had focused on biology instead of physics, he would have likely made similar discoveries in that field as he did in physics. His laws of motion and universal gravitation could have been formulated in a biological context, for example, describing the ways in which cells and organisms adhere to physical laws. He would have likely realized that biology is also an essential physical science, and that the laws of motion are just as universal in biology as in physics. \\ \\
    \textbf{Dynamic Early Exiting:} 
    Isaac Newton had focused on biology instead of physics, he would have developed a different understanding of the natural world. He would have believed that biology was the foundation of all physics, and that it was the only way to understand the universe. He would have also believed that all living organisms had the same underlying laws of physics, rather than the laws of physics being created for humans to understand. He would have also believed that biology was the most important branch of science, and that it was the only way to understand the universe. \\ 
    } 
    
\end{tcolorbox}

These examples show that the responses have high syntactic and semantic similarity. Furthermore, the early exiting responses do not compromise the generation quality.

\subsection{Quality and Inference Cost Analysis at Category Level}

\begin{figure*}[htbp]
\centering
    \begin{subfigure}{0.7\linewidth}
        \includegraphics[width=\linewidth]{Pictures/quality_and_flops/legend_2.pdf}
    \end{subfigure}

    \begin{subfigure}{.19\linewidth}
        \includegraphics[width=\linewidth]{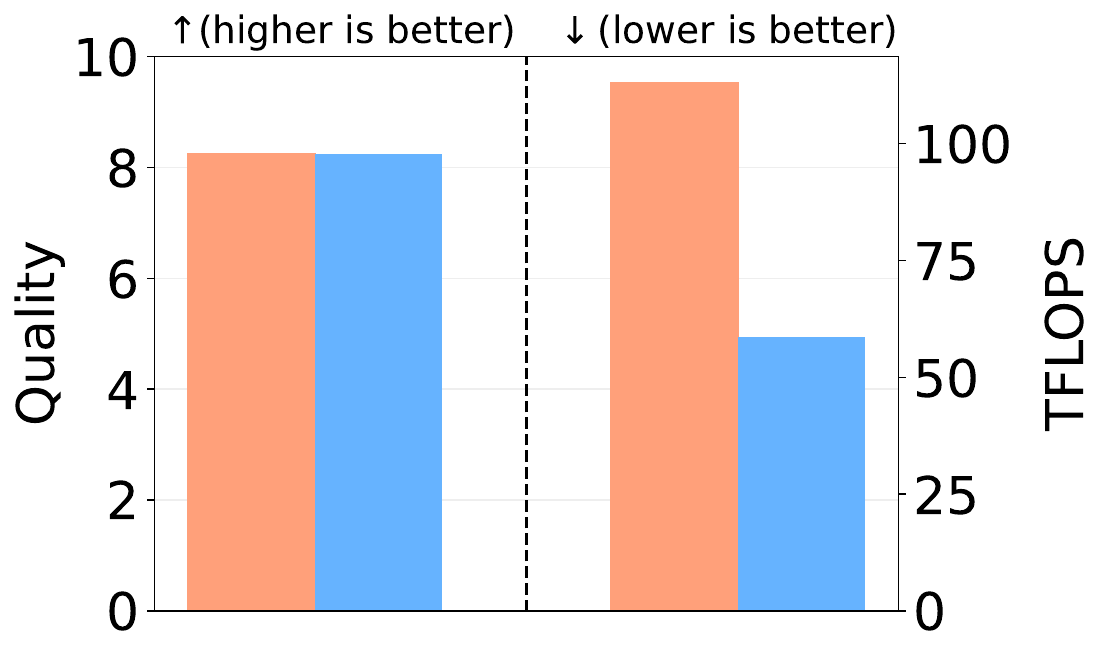}        
        \caption{Generic}        
    \end{subfigure}
    \begin{subfigure}{.19\linewidth}
         \includegraphics[width=\linewidth]{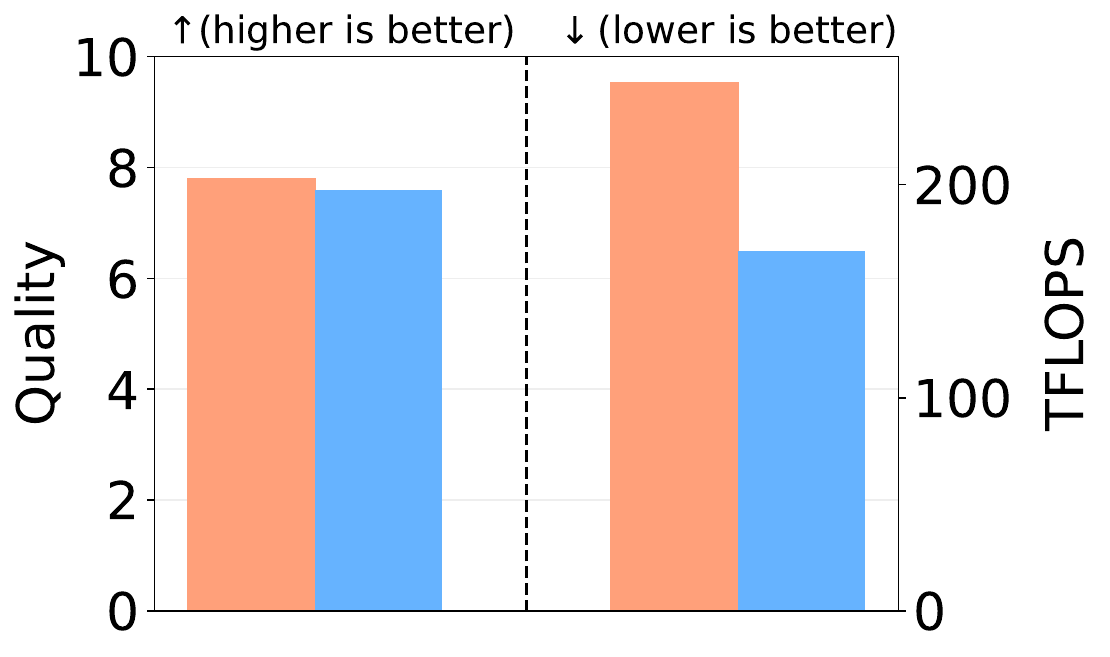}         
         \caption{Knowledge}
    \end{subfigure}    
    \begin{subfigure}{.19\linewidth}
         \includegraphics[width=\linewidth]{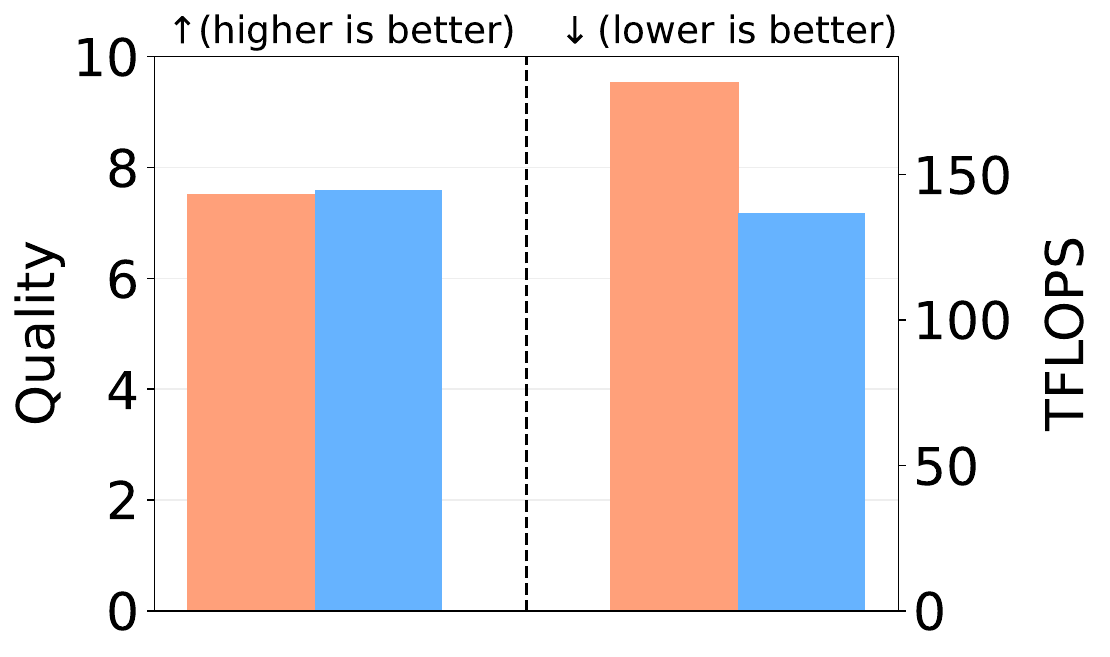}         
         \caption{Roleplay}        
    \end{subfigure}
    \begin{subfigure}{.19\linewidth}        
         \includegraphics[width=\linewidth]{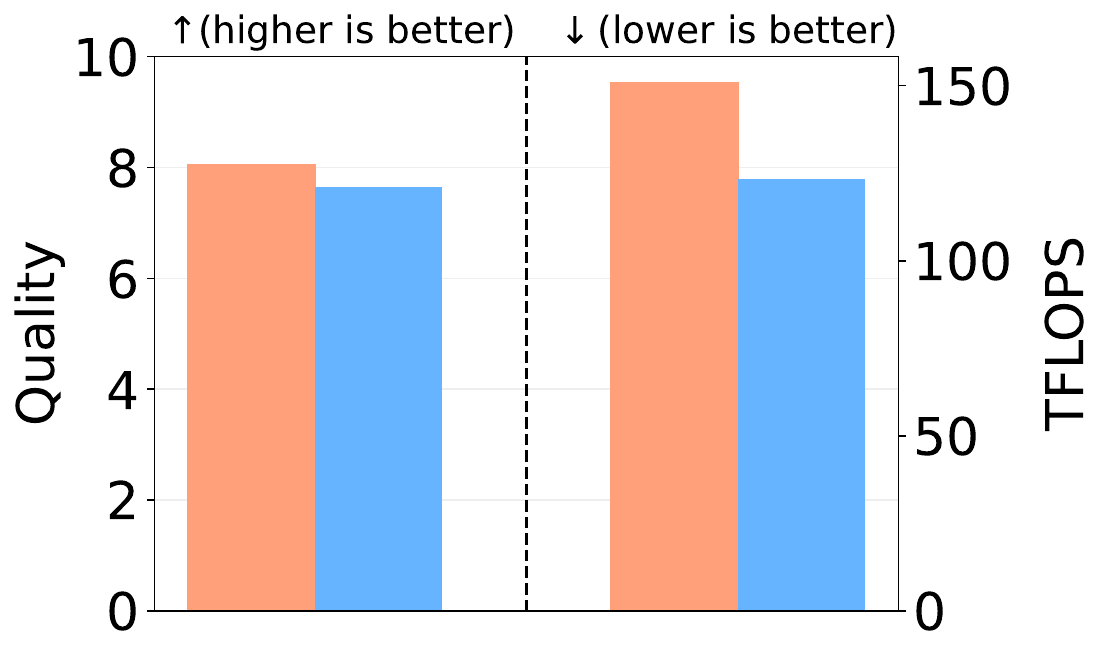}         
         \caption{Commonsense}
    \end{subfigure} 
    \begin{subfigure}{.19\linewidth}        
         \includegraphics[width=\linewidth]{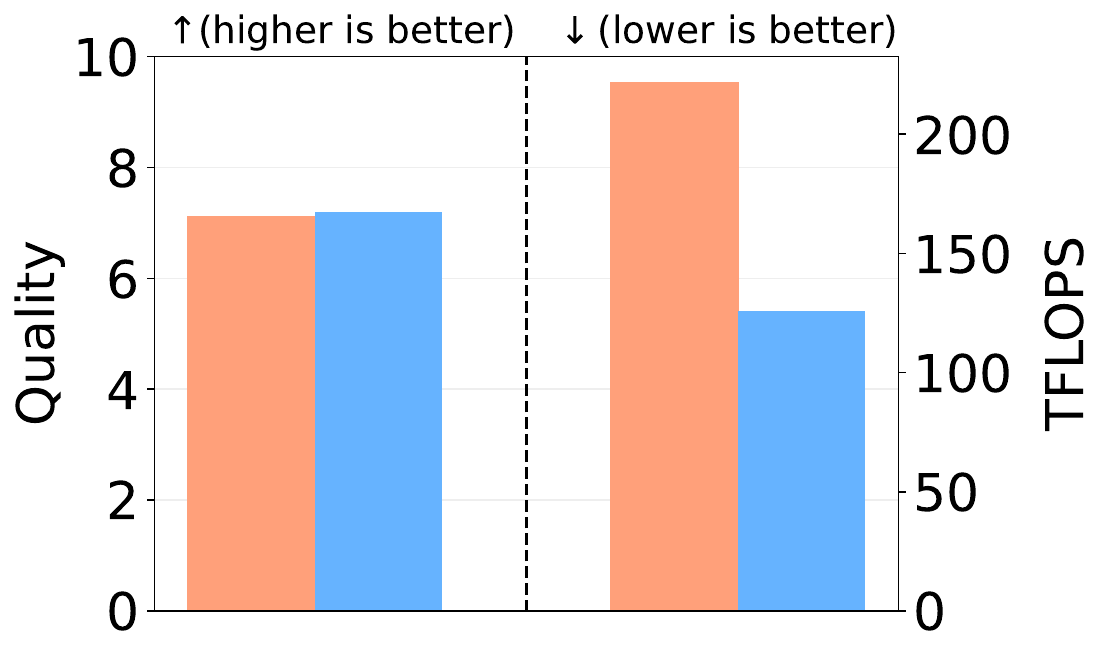}         
         \caption{Fermi}
    \end{subfigure}  
    
    \begin{subfigure}{.19\linewidth}        
         \includegraphics[width=\linewidth]{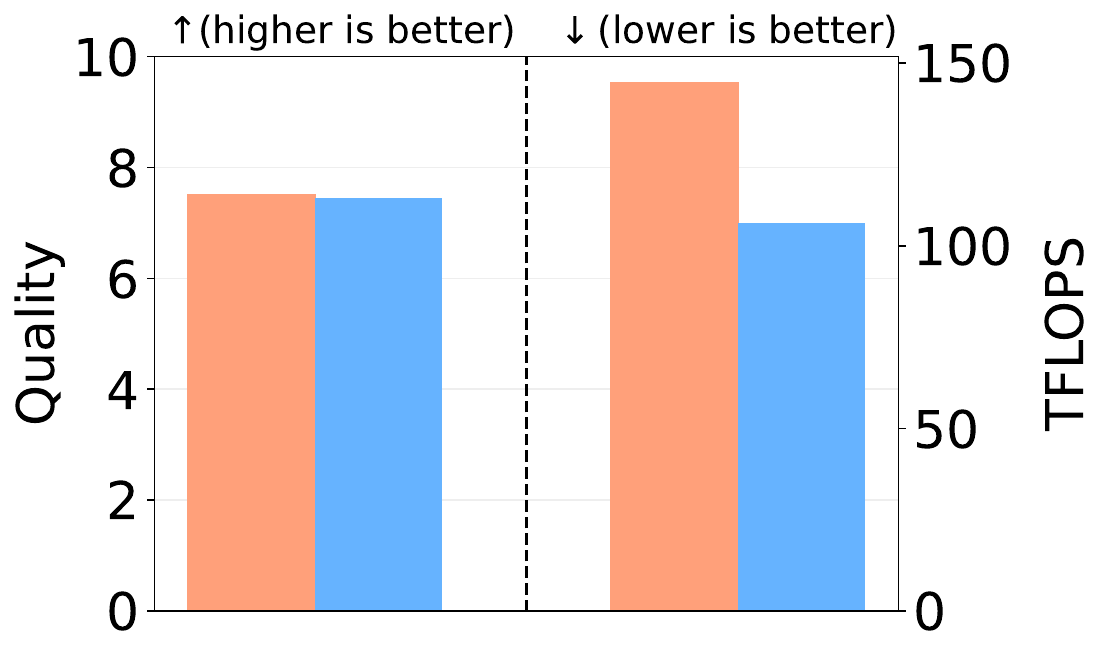}         
         \caption{Counterfactual}
    \end{subfigure}  
    \begin{subfigure}{.19\linewidth}        
         \includegraphics[width=\linewidth]{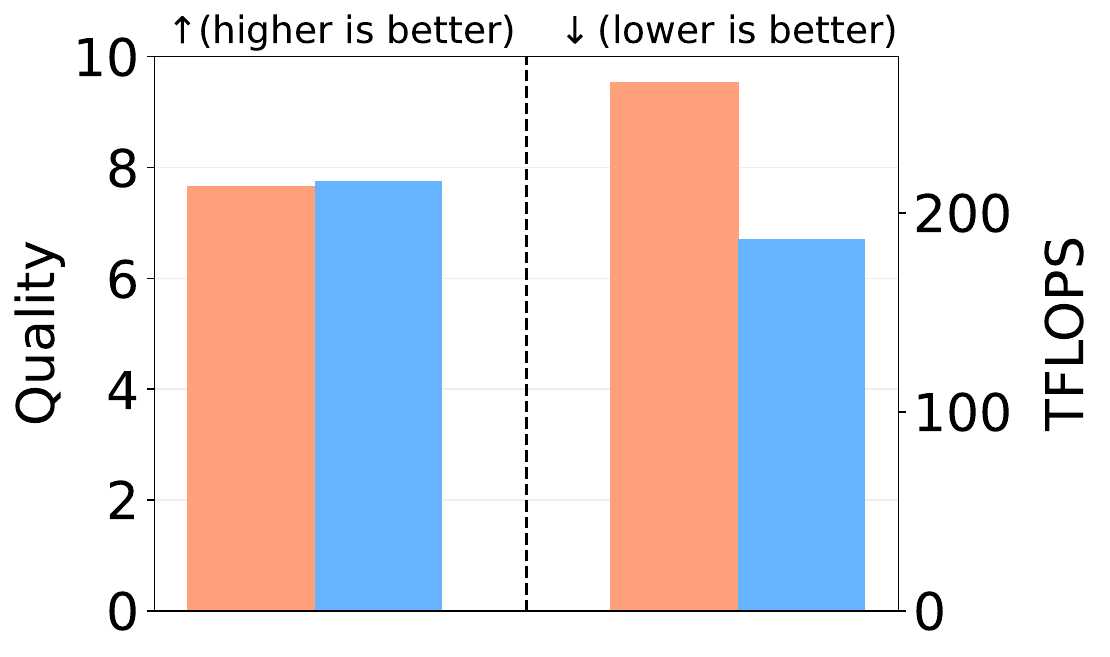}         
         \caption{Writing}
    \end{subfigure}  
    \begin{subfigure}{.19\linewidth}        
         \includegraphics[width=\linewidth]{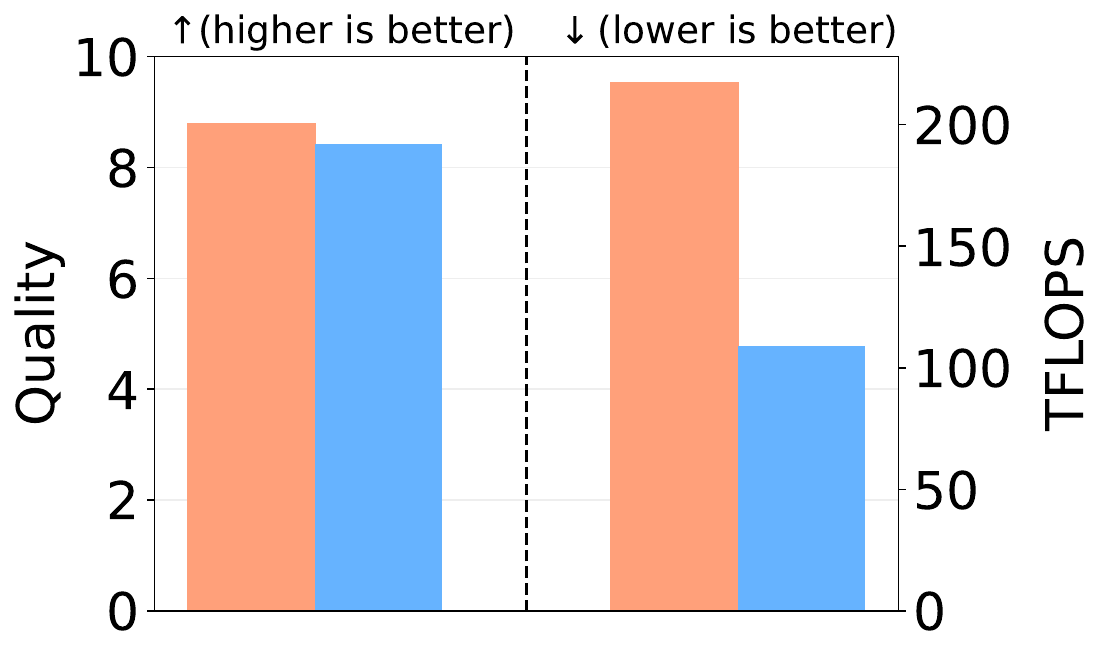}         
         \caption{Coding}
    \end{subfigure} 
    \begin{subfigure}{.19\linewidth}        
         \includegraphics[width=\linewidth]{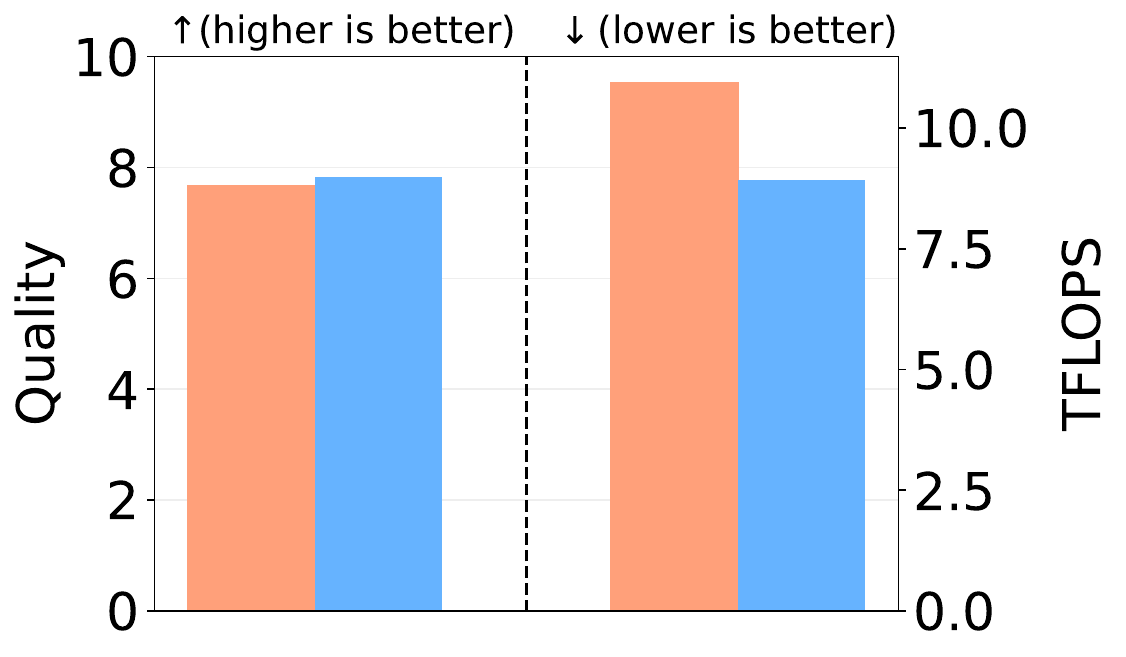}         
         \caption{Math}
    \end{subfigure} 
    
    \caption{Comparing the quality of responses and the inference cost of the standard generation method with our dynamic early exiting method on different categories of the \textbf{Vicuna Test set}. 
    }
    \label{fig:vicuna_category_level_performance}    
\end{figure*}
Vicuna and WizardLM test sets also provide the category corresponding to different test instances. 
To this end, we present category-level quality and inference cost results for these datasets.

\paragraph{Vicuna: }
Figure \ref{fig:vicuna_category_level_performance} compares the quality of responses and the inference cost of the standard generation method (final layer) with our dynamic early exiting method for different categories of Vicuna test set.
On average, it results in cost improvement of $33.39\%$.

\paragraph{WizardLM:}
Figure \ref{fig:wizardLM_category_level_performance} compares the quality of responses and the inference cost of the standard generation method (final layer) with our dynamic early exiting method for different categories of WizardLM test set. 
On average, it results in cost improvement of $36.12\%$.

\begin{figure*}[htbp]
\centering
    \begin{subfigure}{0.7\linewidth}
        \includegraphics[width=\linewidth]{Pictures/quality_and_flops/legend_2.pdf}
    \end{subfigure}

    \begin{subfigure}{.19\linewidth}
        \includegraphics[width=\linewidth]{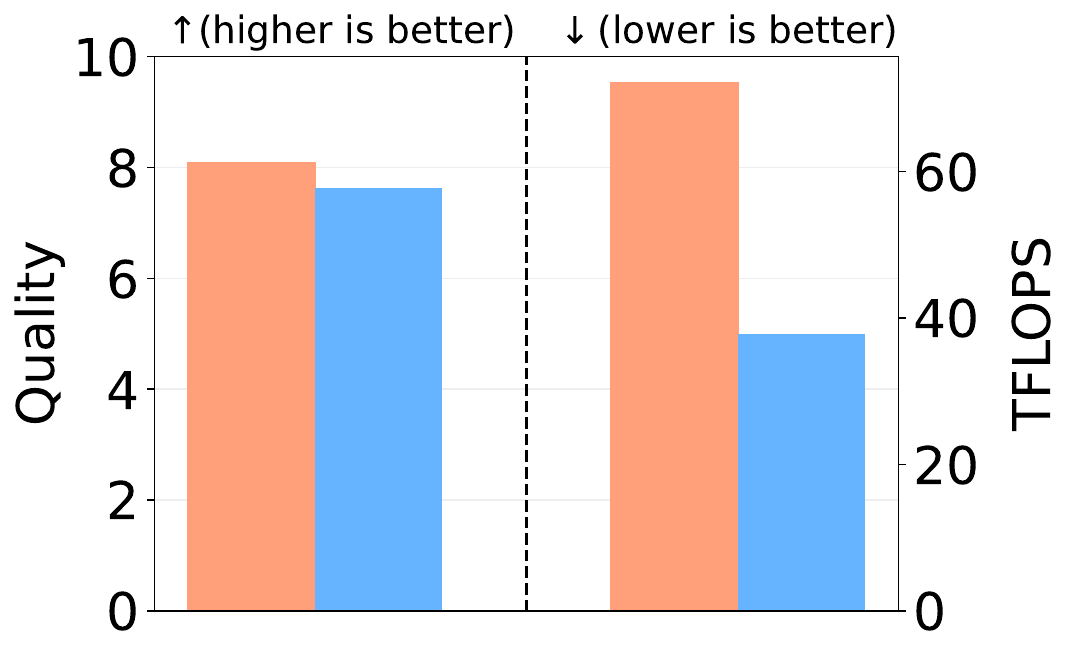}        
        \caption{Math}        
    \end{subfigure}
    \begin{subfigure}{.19\linewidth}
         \includegraphics[width=\linewidth]{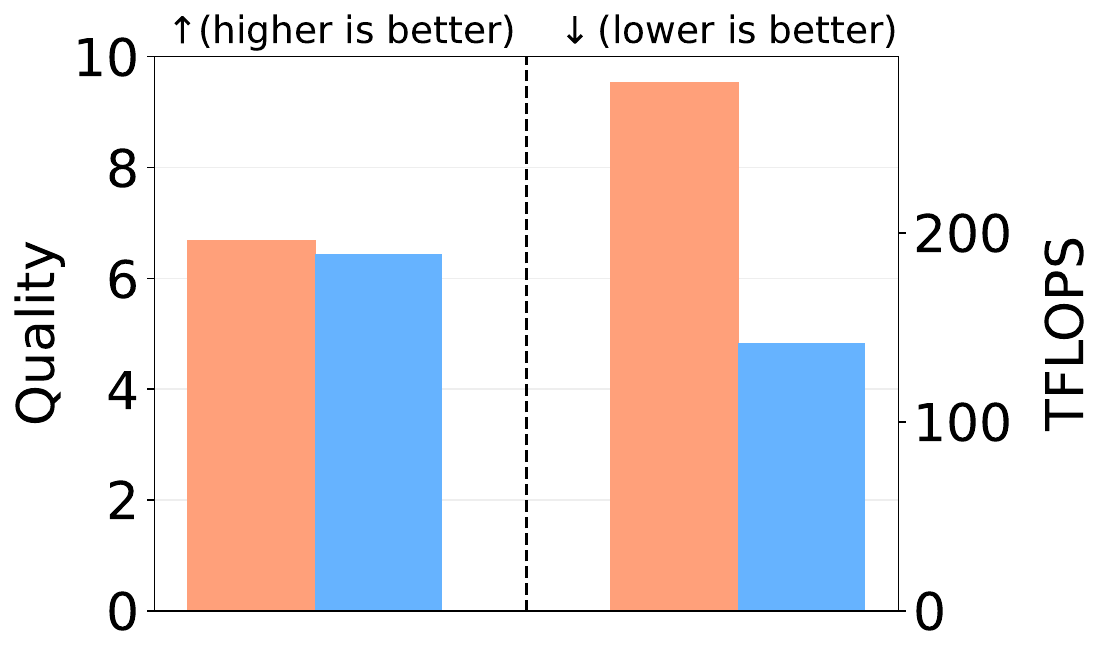}         
         \caption{Code Generation}
    \end{subfigure}    
    \begin{subfigure}{.19\linewidth}
         \includegraphics[width=\linewidth]{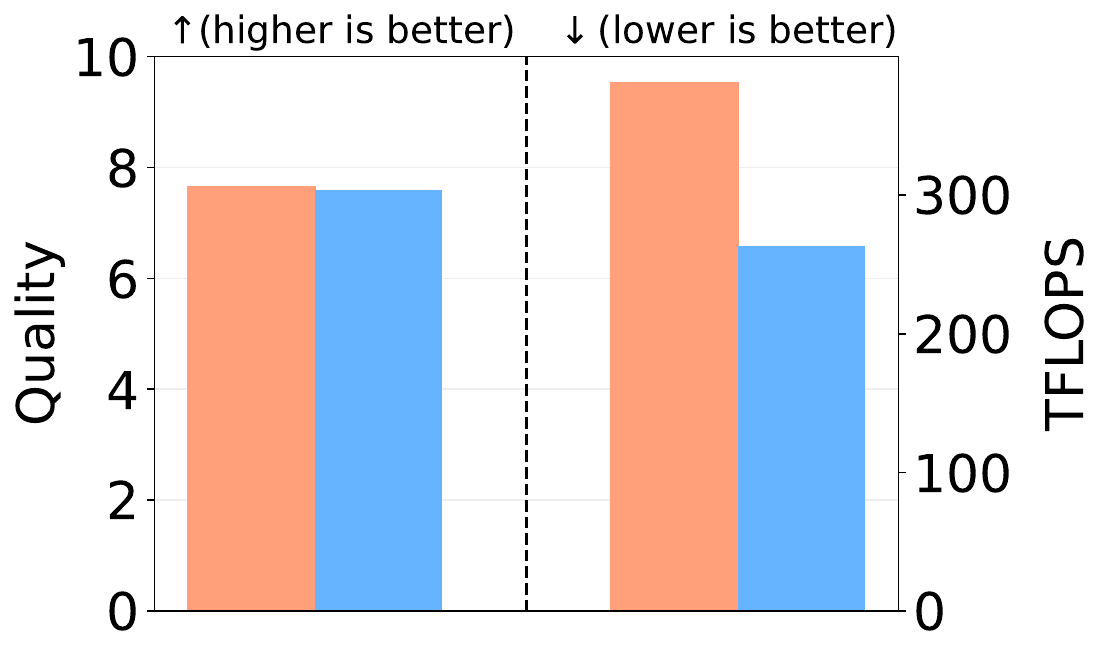}         
         \caption{Writing}        
    \end{subfigure}
    \begin{subfigure}{.19\linewidth}        
         \includegraphics[width=\linewidth]{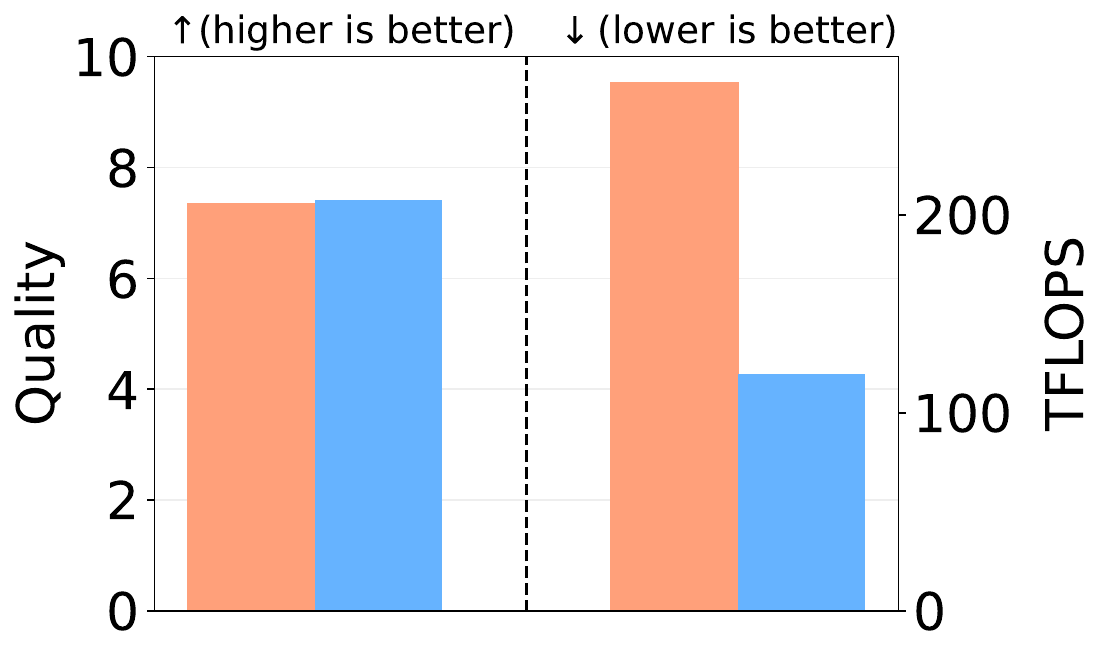}         
         \caption{Reasoning}
    \end{subfigure} 
    \begin{subfigure}{.19\linewidth}        
         \includegraphics[width=\linewidth]{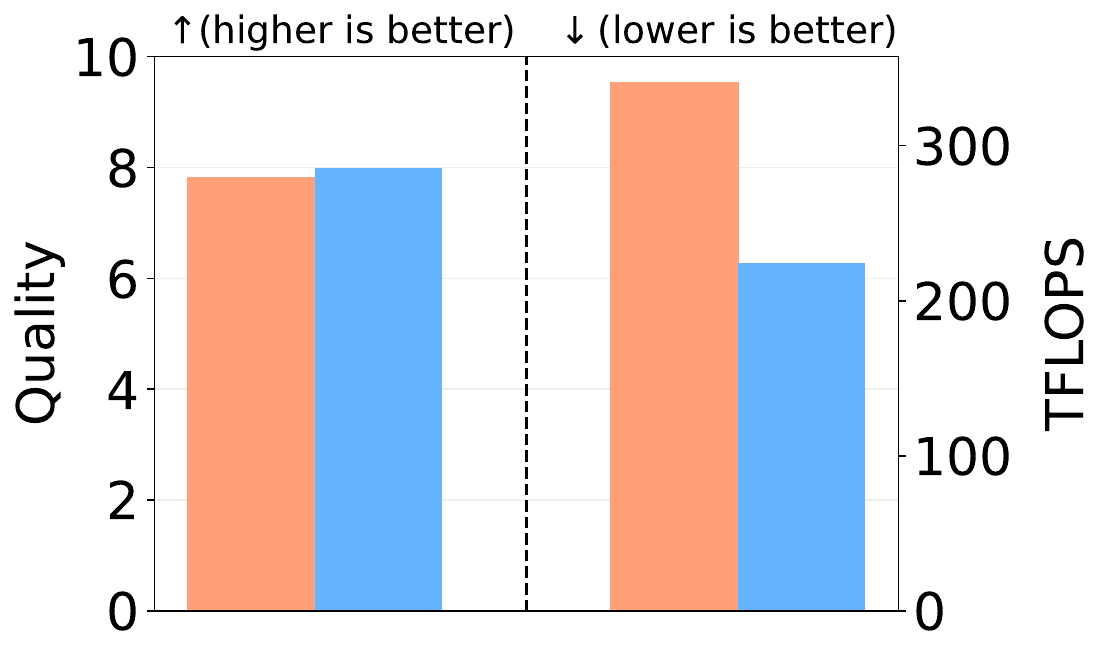}         
         \caption{Computer Sc. }
    \end{subfigure}  
    
    \begin{subfigure}{.19\linewidth}        
         \includegraphics[width=\linewidth]{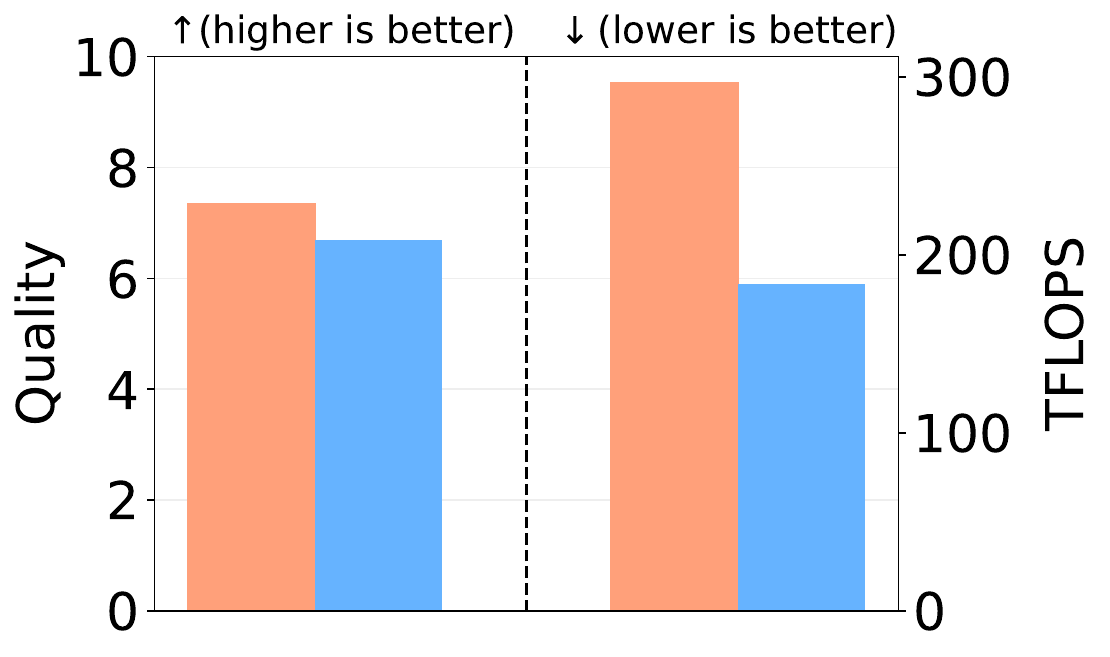}         
         \caption{Code Debug}
    \end{subfigure}  
    \begin{subfigure}{.19\linewidth}        
         \includegraphics[width=\linewidth]{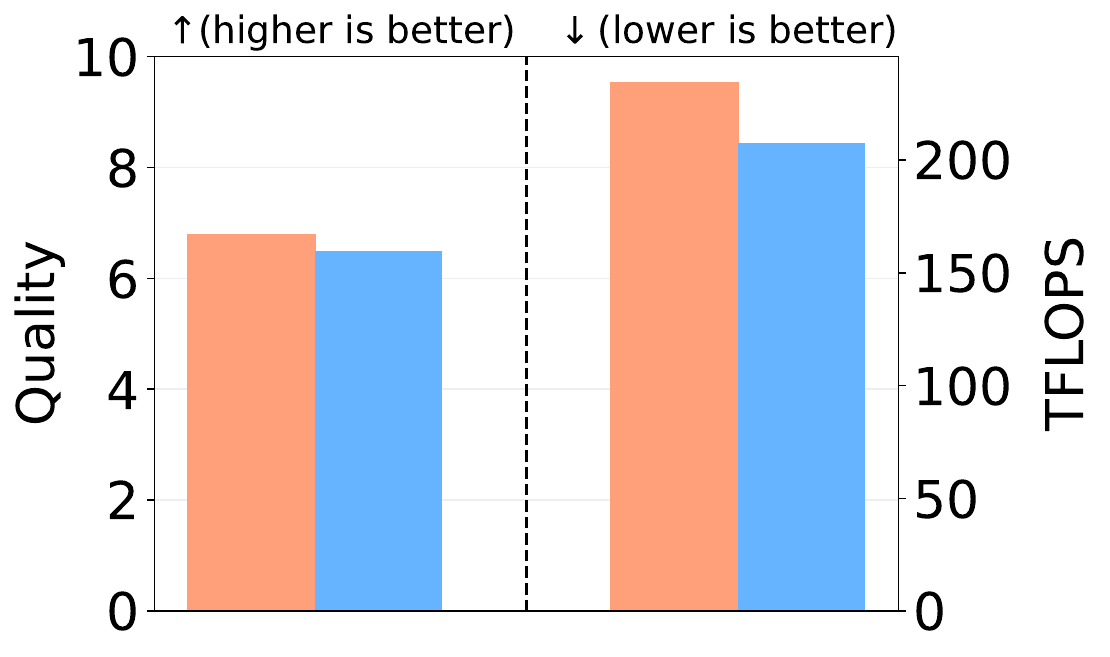}         
         \caption{Complex Format }
    \end{subfigure}  
    \begin{subfigure}{.19\linewidth}        
         \includegraphics[width=\linewidth]{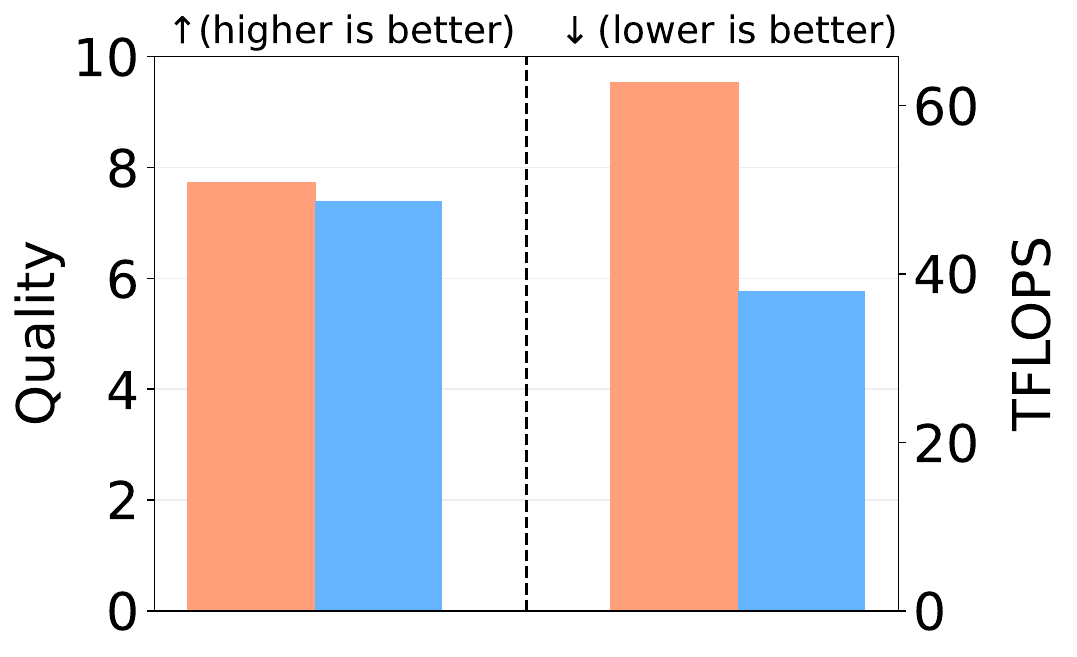}         
         \caption{Common-Sense}
    \end{subfigure} 
    \begin{subfigure}{.19\linewidth}        
         \includegraphics[width=\linewidth]{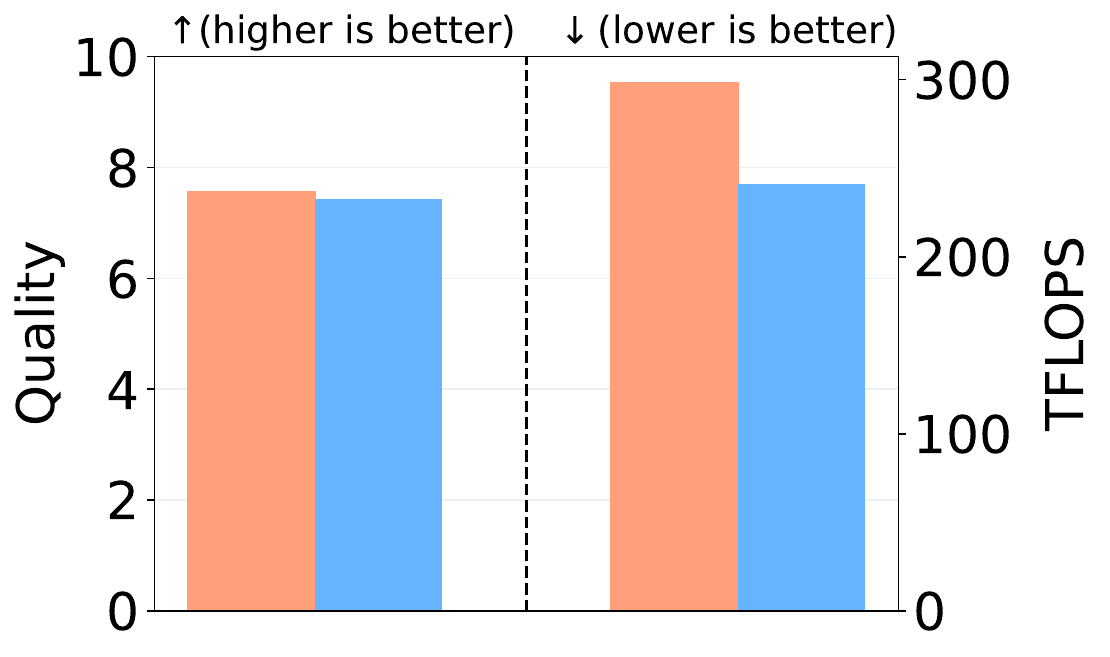}         
         \caption{Counterfactual}
    \end{subfigure} 
    \begin{subfigure}{.19\linewidth}        
         \includegraphics[width=\linewidth]{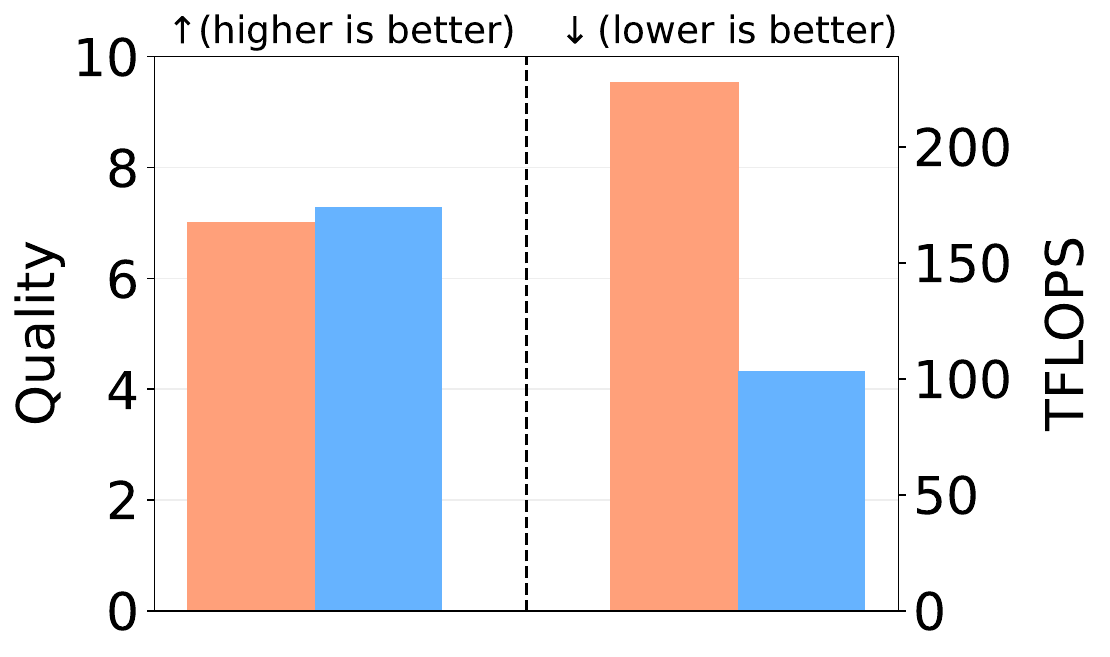}         
         \caption{Multilingual}
    \end{subfigure}
    
    \caption{Comparing the quality of responses and the inference cost of the standard generation method with our dynamic early exiting method on different categories of the \textbf{WizardLM Test set}. 
    }
    \label{fig:wizardLM_category_level_performance}    
\end{figure*}

\subsection{Relationship Between Token Prediction Confidence and Percentage Alignment of the Intermediate Layers for the Model Tuned with Instruction Tuning (IT)}
\label{sec_confidence_alignment_IT}

\begin{figure*}[htbp]
\centering
    \begin{subfigure}{0.5\linewidth}
        \includegraphics[width=\linewidth]{Pictures/RQ2/legend.pdf}
    \end{subfigure}

    \begin{subfigure}{.24\linewidth}
        \includegraphics[width=\linewidth]{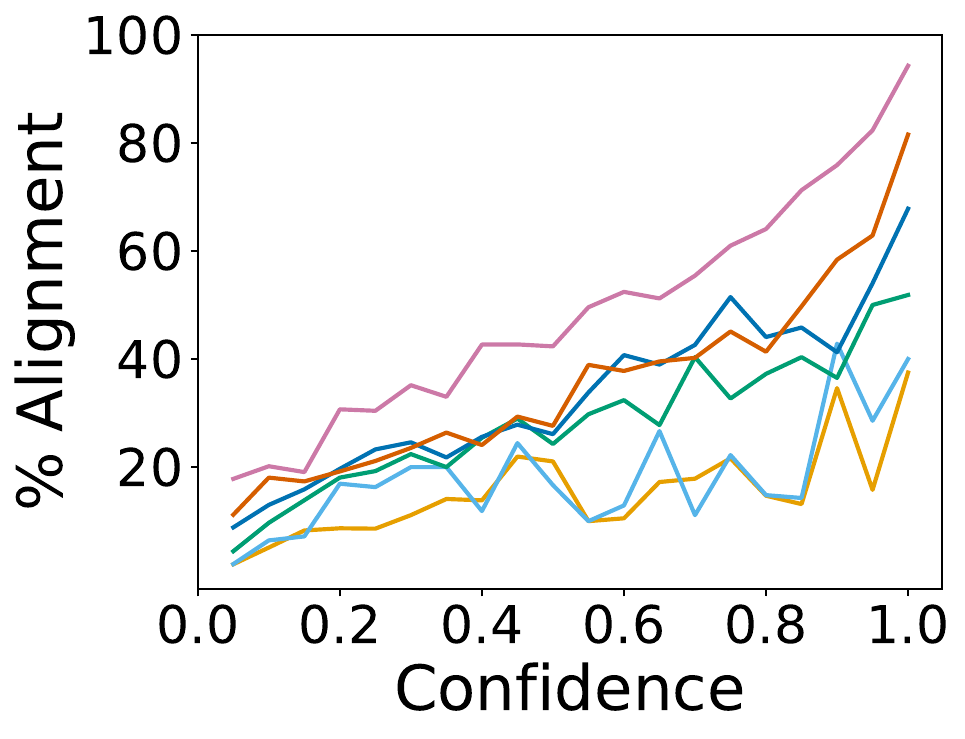}        
        \caption{Vicuna}
    \end{subfigure}
    \begin{subfigure}{.24\linewidth}
         \includegraphics[width=\linewidth]{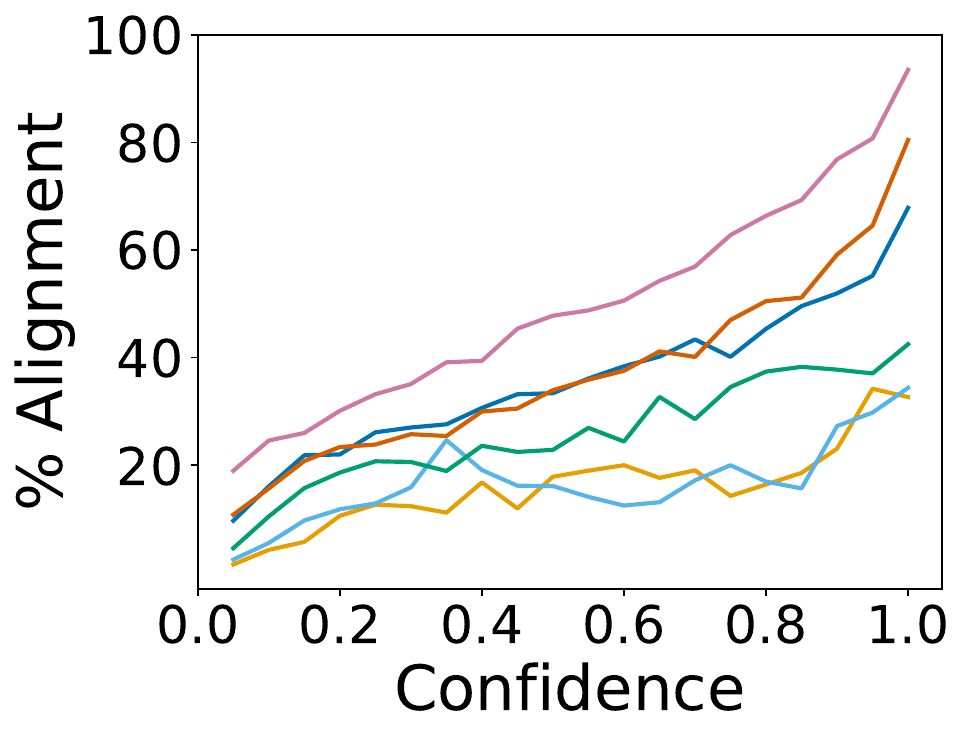}         
         \caption{Koala}
    \end{subfigure}
    \begin{subfigure}{.24\linewidth}        
         \includegraphics[width=\linewidth]{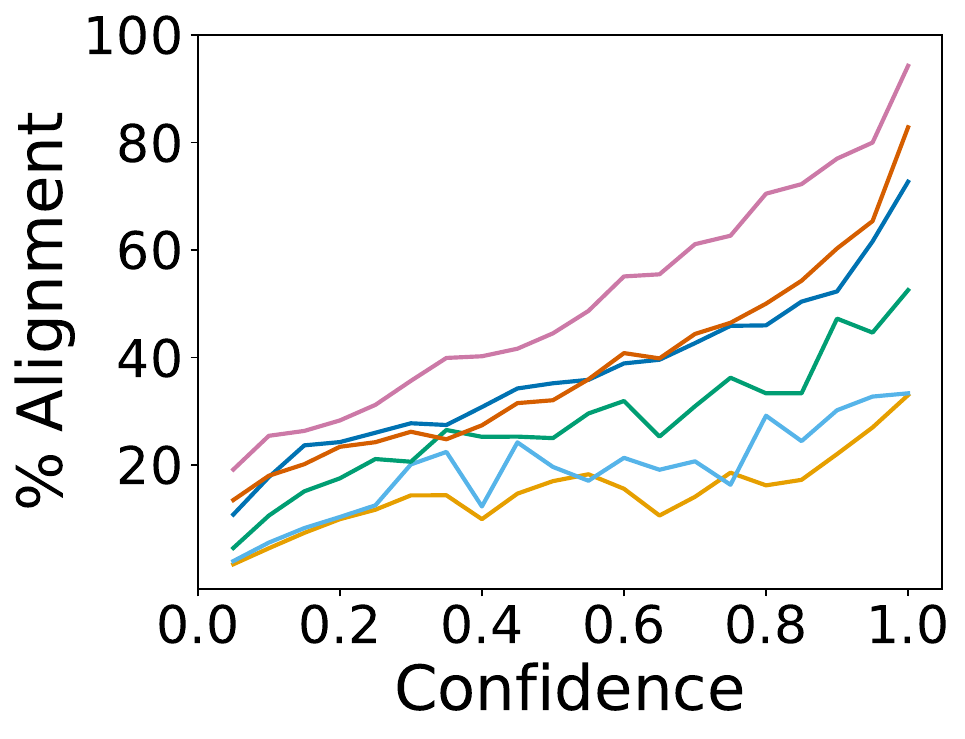}       
         \caption{WizardLM}
    \end{subfigure}    
    \begin{subfigure}{.24\linewidth}
         \includegraphics[width=\linewidth]{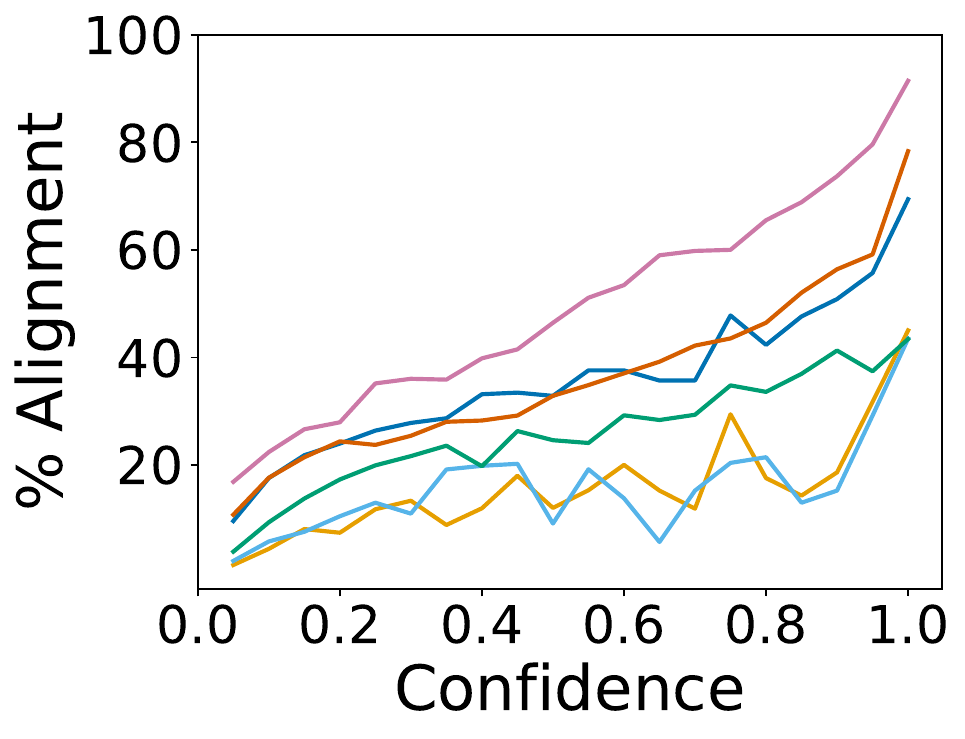}  
         \caption{Self-Instruct}
    \end{subfigure}

    \caption{Demonstrating relationship between token prediction confidence of the intermediate layers and the percentage alignment with the token prediction of the final layer for model tuned with IT.    
    }
    \label{fig:confidence_alignment_IT}    
\end{figure*}

Figure \ref{fig:confidence_alignment_IT} shows the relationship between the token prediction confidence of the intermediate layers and the percentage alignment with the token prediction of the final layer for standard instruction tuning (IT).
It shows that the confidence is not well correlated with the percentage alignment. 
However, in IT with LITE (Figure \ref{fig:confidence_alignment}), the intermediate layers' token prediction probabilities provide a strong signal of alignment.

\subsection{Dynamic Confidence-Based Early Exiting with Aggressive Confidence Thresholds}
\label{appendix_configuration}

We also experiment with aggressive confidence thresholds. Specifically, we use the following confidence thresholds:
Layer 8: $0.85$, 
Layer 12: $0.85$, 
Layer 16: $0.8$, 
Layer 20: $0.8$,
Layer 24: $0.7$, and
Layer 28: $0.6$.
These thresholds are lower than those used in the main paper.
Figure \ref{fig:early_exiting_results_configuration_2} shows the quality and cost comparisons.
It leads to larger cost improvements (of $49.92\%$) though it slightly drops the quality of generation (by $5.34\%$).

\begin{figure*}[htbp]
\centering
    \begin{subfigure}{0.7\linewidth}
        \includegraphics[width=\linewidth]{Pictures/quality_and_flops/legend_2.pdf}
    \end{subfigure}

    \begin{subfigure}{.24\linewidth}
        \includegraphics[width=\linewidth]{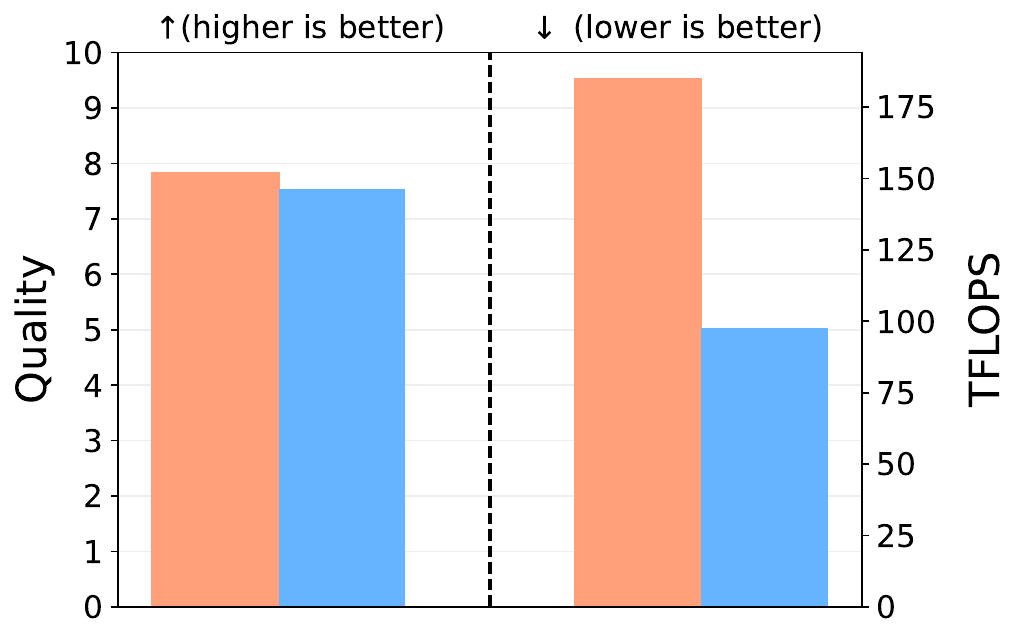}        
        \caption{Vicuna}        
    \end{subfigure}
    \begin{subfigure}{.24\linewidth}
         \includegraphics[width=\linewidth]{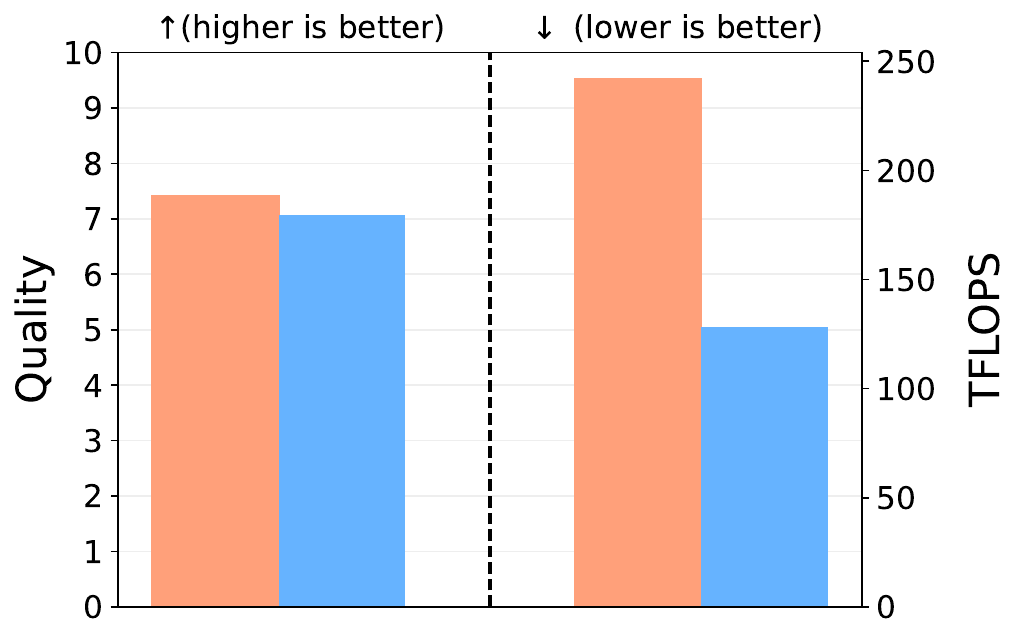}         
         \caption{Koala}
    \end{subfigure}    
    \begin{subfigure}{.24\linewidth}        
         \includegraphics[width=\linewidth]{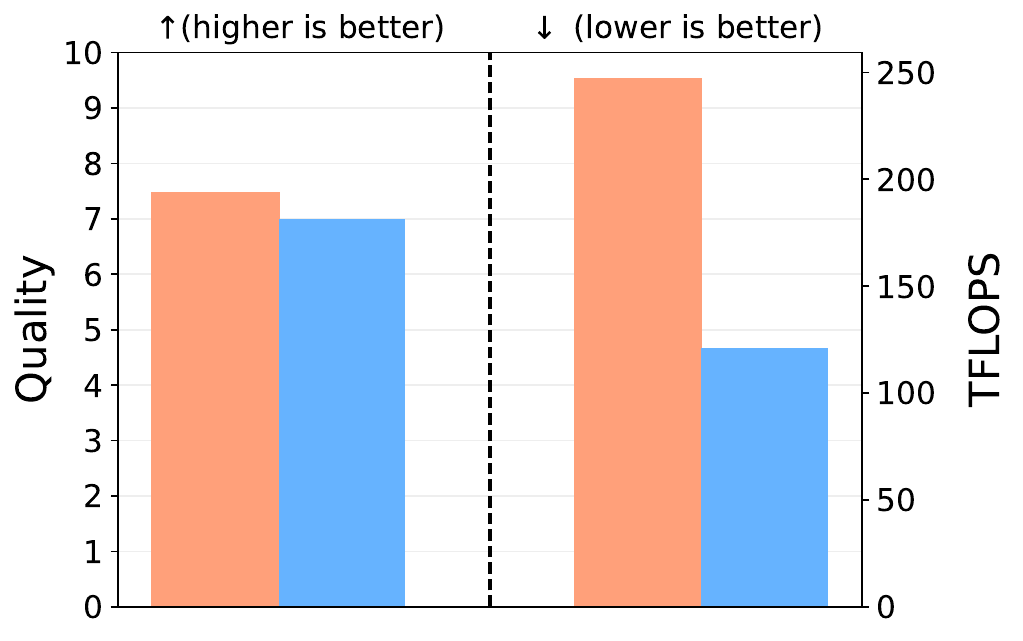}         
         \caption{WizardLM}
    \end{subfigure}    
    \begin{subfigure}{.24\linewidth}
         \includegraphics[width=\linewidth]{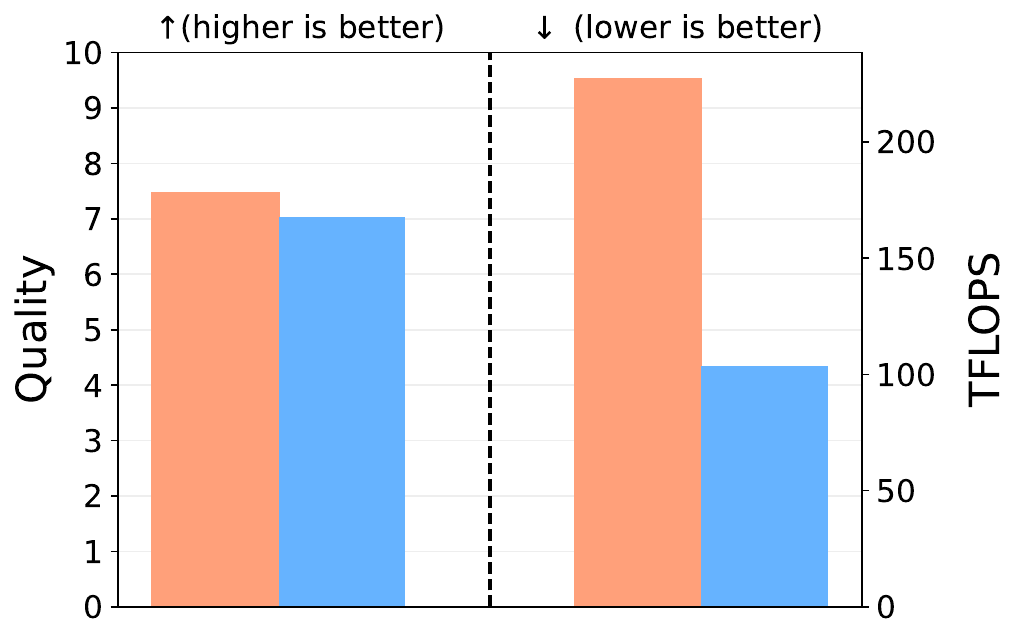}         
         \caption{Self-Instruct}        
    \end{subfigure}
    
    \caption{Comparing the quality of responses (evaluated using the Claude model) and the inference cost (measured in FLOPs) of the standard generation method from the final layer with our dynamic early exiting method. 
    Confidence Thresholds: Layer 8: $0.85$, 
    Layer 12: $0.85$, 
    Layer 16: $0.8$, 
    Layer 20: $0.8$,
    Layer 24: $0.7$, and
    Layer 28: $0.6$.
    This \textbf{aggressive configuration} results in larger cost improvements of $49.93\%$ but results in a slight degradation in the generation quality.
    }
    \label{fig:early_exiting_results_configuration_2}    
\end{figure*}

\subsection{Results for 13B Model}
\label{sec_13b_results}

For the 13B model we use the following confidence thresholds:
Layer 8: $0.95$, 
Layer 12: $0.95$, 
Layer 16: $0.9$, 
Layer 20: $0.9$,
Layer 24: $0.8$, 
Layer 28: $0.7$,
Layer 32: $0.7$, and
Layer 36: $0.65$,

\begin{table}[t]
    \centering
    {
    \begin{tabular}{@{}cc@{}}
        \toprule
         \textbf{Test Dataset} & \textbf{Cost Improvement (\%)}\\
         
        \midrule
        
        Vicuna & 43.60 \% \\
        Koala & 45.62 \% \\
        WizardLM & 50.84 \%\\
        Self Instruct & 45.35 \% \\
        
    \bottomrule
    \end{tabular}    
    }
    \caption{
    Percentage improvements in the inference cost (measured in FLOPs) with dynamic early exiting for the 13B model. On average, it results in an improvement of 46.35\%.
    }

    \label{tab:percentage_improvements_13B}
\end{table}

Table \ref{tab:percentage_improvements_13B} shows the cost improvements resulting from dynamic early exiting from the 13B model on each test dataset.
On average, it results in $46.35\%$ cost improvement.

\end{document}